\documentclass[lettersize,journal]{IEEEtran}

\makeatletter
\IfFileExists{hyperref.sty}{}{\def\href#1#2{#2}}
\makeatother

\usepackage{amsmath,amsfonts}
\usepackage{algorithmic}
\usepackage{algorithm}

\usepackage{array}
\usepackage[caption=false,font=normalsize,labelfont=sf,textfont=sf]{subfig}
\usepackage{textcomp}
\usepackage{stfloats}
\usepackage{url}
\usepackage{verbatim}
\usepackage{graphicx}
\usepackage{booktabs}
\usepackage{tabularx}
\usepackage{multirow}
\usepackage{cite}
\usepackage[labelfont=bf]{caption}
\usepackage{hyperref}
\hypersetup{
    colorlinks=true,
    linkcolor=blue,
    citecolor=blue,
    urlcolor=blue
}

\begin{document}

\title{ Predicting Time Pressure of Powered Two-Wheeler Riders for Proactive Safety Interventions}

\author{
    Sumit S. Shevtekar\textsuperscript{1},
    Chandresh K. Maurya\textsuperscript{1},
    Gourab Sil\textsuperscript{2}
    \thanks{Corresponding author: Sumit Shevtekar, email: phd2401101010@iiti.ac.in}%
    \thanks{\textsuperscript{1}Department of Computer Science and Engineering, Indian Institute of Technology Indore, India. Email: chandresh@iiti.ac.in}%
    \thanks{\textsuperscript{2}Department of Civil Engineering, Indian Institute of Technology Indore, India. Email: gourabsil@iiti.ac.in}%

}

% The paper headers
\markboth{}%
{Shell \MakeLowercase{\textit{et al.}}: A Sample Article Using IEEEtran.cls for IEEE Journals}

\IEEEpubid{}

\maketitle
\IEEEpubidadjcol

\begin{abstract}

Time pressure critically influences risky maneuvers and crash proneness among powered two-wheeler riders, yet its prediction remains underexplored in intelligent transportation systems. To address this gap, we propose MotoTimePressure (MTPS), a deep learning model combining convolutional preprocessing, dual-stage temporal attention, and Squeeze-and-Excitation feature recalibration, achieving 91.53\% accuracy and 98.93\% ROC AUC, outperforming six baselines, with only 172K parameters, 0.66 MB model size, and 0.21 ms inference on CPU. To validate and benchmark MTPS, we present a dataset of 129,209 feature windows from 153 simulator sessions by 51 experienced male PTW riders under No, Low, and High Time Pressure conditions. Each sequence captures 63 features spanning vehicle kinematics, control inputs, behavioral violations, and environmental context. Our empirical analysis shows High Time Pressure induces 48\% higher speeds, 36.4\% greater speed variability, 58\% more risky turns at intersections, 36\% more sudden braking, and 50\% higher rear brake forces versus No Time Pressure. Since time pressure cannot be directly measured in real time, we demonstrate its utility in collision prediction and threshold determination. Using MTPS-predicted time pressure as a feature improves collision risk accuracy for both Informer (91.25\% to 93.51\%) and TimesNet (92.10\% to 93.90\%), approaching oracle performance (93.72\% and 94.06\%, respectively). Thresholded time pressure states capture rider cognitive stress and enable proactive ITS interventions, including adaptive alerts, haptic feedback, V2I signaling, and speed guidance, supporting safer two-wheeler mobility under the Safe System Approach.

\end{abstract}


\section{Introduction}

\IEEEPARstart{P}{owered} two-wheelers (PTWs) are among the most accessible and widely used transport modes, particularly in developing countries. However, they carry elevated crash risks due to exposure to rider and balance control demands~\cite{10122165}. The severity of the crash is strongly influenced by overspeeding, traffic dynamics, rider behavior, and road conditions~\cite{Seefong2024,7328721,9625971}, with human error and risky maneuvers being the leading causes~\cite{pavlidis2016dissecting,6899632,CHOUHAN20241378}. PTWs account for 21\% of global road fatalities rising to 38\% of deaths and 51\% of injuries in Low- and Middle-Income Countries (LMICs)~\cite{WHO2023}. Almost half of traffic deaths involve vulnerable users such as pedestrians, cyclists, and PTW riders~\cite{WHO2023}. Excess speed alone contributes to 34\% PTW fatalities, with riders traveling up to 2.3 times faster than car drivers~\cite{Sharma21042025}.

\begin{figure}
    \centering
\includegraphics[width=0.48\textwidth]{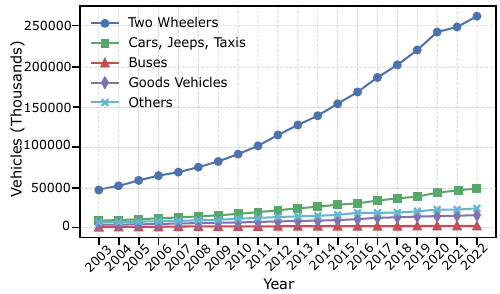}
    \caption{Registered motor vehicles in India by category (2003–2022)~\cite{MORTH202425}.}
    \label{fig:vehicle_trend}
\end{figure}

According to India’s Ministry of Road Transport and Highways (MoRTH), PTWs are the most prevalent and economical mode of transport in LMICs. Between 2003 and 2022, India’s registered vehicle population grew to 354 million, of which PTWs constituted 263 million (74.4\%). This segment grew at a CAGR of 8.6\%, surpassing buses (2.48\%) and goods vehicles (7.3\%). Category-wise registration trends are shown in Fig.~\ref{fig:vehicle_trend}, underscoring the growing dependence on PTWs and the urgent need for targeted safety interventions~\cite{MORTH202425}.

PTWs also account for the largest share of traffic fatalities in India, accounting for 44.5\% of deaths in 2022, followed by pedestrians (19.5\%), cars/LMVs (12.5\%), trucks/lorries (6.3\%), autorickshaws (3.9\%), bicycles (2.9\%), buses (2.4\%), and others (8.0\%)~\cite{MORTH202324,MORTH202122,MORTH202223}. According to official MoRTH reports on road accidents in india, road fatalities in India are predominantly male, ranging from 85.2\% to 87.3\% between 2019 and 2023~\cite{morth2019,morth2020,morth2021,morth2022,morth2023}, with most fatalities occurring in the 18–45 age group. These statistics show the high risk associated with male PTW riding behaviour. Driving errors are the main cause of crash globally, especially for PTWs, which face high instability and dynamic control challenges~\cite{8294048,BHALLA201583}. Field studies of PTW behavior have shown significant variability in acceleration and braking patterns~\cite{Mondal21042023}, emphasizing the inherent dynamic control challenges these vehicles face in real-world traffic. Simulator studies further reveal context-dependent rider behavior in mixed traffic~\cite{10073653}, underscoring PTWs vulnerability and the need for targeted safety interventions.

A critical yet underexplored factor in PTW safety is time pressure (TP), reflecting perceived urgency or limited time to complete riding tasks~\cite{PAWAR2022105582}. For example, many delivery PTW riders in India are subjected to TPs~\cite{fortune2025quickcommerce,thehindu2022delivery,jakartapost2022delivery}. Elevated TP impairs cognition and motor control, driving overspeeding, braking inconsistencies, delayed hazard response, and higher crash risk~\cite{PAWAR2022105582,Gupta25112024,Sharma21042025}. Naturalistic studies link risky behavior to speed, maneuvering, and context, while stress further amplifies variability and control errors~\cite{pavlidis2016dissecting}. Field evidence from Thailand also confirms that widespread speeding significantly raises collision risk~\cite{Seefong2024}. However, existing studies have only examined TP as a cause of risky behavior, not as a predictable cognitive state that can be detected in real-time.

Pawar et al.~\cite{PAWAR2022105582} showed that TP induces risky maneuvers, but directly measuring TP in real-world PTW riding remains practically challenging. We hypothesize that TP can be predicted from PTW's data. To test this hypothesis, we first collected a first-of-its-kind supervised dataset of 129,209 feature windows extracted via sliding window segmentation from 153 rides by 51 participants under TP conditions (NTP, LTP, HTP). Using this dataset, we propose MotoTimePressure (MTPS), a novel deep learning model specifically designed for rider TP state prediction — the first predictive model for cognitive state detection in PTW riders, addressing a fundamental limitation in current ITS. To demonstrate the utility of predicted TP, we train an Informer~\cite{zhou2021informer} and TimesNet~\cite{wu2023timesnettemporal2dvariationmodeling} model using ground truth (GT) TP and MTPS-predicted TP features. Our results show that predicted TP improves collision risk accuracy from 91.25\% to 93.51\% for Informer and from 92.10\% to 93.90\% for TimesNet, approaching oracle performance (93.72\% and 94.06\%, respectively). The key contributions of this work are summarized as follows:

\begin{enumerate}
\item The First Labeled Dataset for PTW Rider Cognitive State: To the best of our knowledge, we present the first high-resolution PTW simulator dataset comprising 129,209 feature windows extracted via sliding window segmentation from 153 rides by 51 participants under three TP conditions (NTP, LTP, HTP). 
  
\item Systematic Behavioral Degradation Under TP: TP systematically degrades riding control and creates compound collision risks. Under HTP, riders exhibit 48\% higher speeds, 36.4\% greater variability, 604\% increased front brake, 50\% higher rear brake, 35\% increased gear usage, 58\% more risky turns, 36\% more sudden braking, and 67\% worse clutch control. Despite natural compensation, human adaptation fails — HTP still causes 21.3\% worse control variability and 10.2\% more violations, following: TP $\rightarrow$ degraded control $\rightarrow$ insufficient correction $\rightarrow$ persistent risk.

\item Predictive Model for Time Pressure: We propose MTPS, the first architecture specifically designed for rider TP state prediction. MTPS introduces a novel integration of convolutional preprocessing, dual temporal attention blocks, and hierarchical feature fusion to capture multi-scale dependencies in riding behavior, achieving 91.53\% accuracy and 98.93\% ROC AUC, outperforming six ML and DL baselines including TST, SHGN, and Informer. Moreover, MTPS achieves 0.21 ms inference on CPU with only 172K parameters and 0.66 MB size, enabling real-time proactive safety on edge devices.
    
\item A New Paradigm for Proactive Safety: Incorporating MTPS-predicted TP into crash risk predictors improves accuracy from 91.25\% to 93.51\% for Informer and from 92.10\% to 93.90\% for TimesNet, approaching oracle performance (93.72\% and 94.06\%, respectively). This demonstrates that TP provides cognitive state information beyond kinematics, enabling pre-crash intervention.

\end{enumerate}

\section{Literature Review}
To establish the foundation for our work, this section reviews existing research on TP in driving and riding contexts, gap-acceptance behavior, PTW-specific vulnerabilities, simulator validity, and machine learning approaches to risk modeling.

\subsection{Time Pressure and Driving Behaviour}

Driving under TP can reduce the driver’s focus, which can lead to irrational decisions and a higher risk of collision~\cite{PAWAR2022105582, Gupta25112024}. Drivers operating under TP exhibit higher speeds, altered driving behaviour, shorter accepted gaps, increased cognitive workload, faster decision cycles, and higher crash risk~\cite{PAWAR20221, PAWAR2022105582, Gupta25112024, Sharma21042025, fortune2025quickcommerce, thehindu2022delivery, jakartapost2022delivery}. Nishant et al.~\cite{PAWAR202229} reported that TP significantly increases longitudinal acceleration, braking force variability, and steering fluctuations in a fixed-base simulator, reinforcing that TP elicits statistically significant shifts toward risk-oriented control behaviour. 

TP is fundamentally a perception of urgency or limited time to complete the driving task. This perceived urgency narrows the attentional bandwidth and accelerates cognitive processing, often forcing drivers and riders to make rapid decisions with reduced situational awareness~\cite{Gupta25112024, Sharma21042025, PAWAR202229}. Although early work primarily focused on four-wheelers~\cite{pavlidis2016dissecting,Gupta25112024}, recent PTW research reveals similar but more sensitive behavioural deterioration under TP, including inconsistent braking, excessive speeding, unstable motor control, and delayed hazard response~\cite{Sharma21042025, PAWAR2022105582}. These PTW-specific vulnerabilities are important because PTWs exhibit inherently higher crash risk and lower protection levels. Simulator-based studies further show that TP increases crash likelihood across multiple riding tasks. In complex maneuvers such as intersections, \textit{Pawar \& Velaga}~\cite{PAWAR2022105582} reported that TP increases cognitive urgency, speed and reduces caution. By narrowing attention and biasing riders toward fast, risk-oriented strategies, TP directly reduces safety margins and increases the probability of unsafe decisions.

\subsection{Time Pressure and Gap-Acceptance Behaviour}

Research on un-signalized intersections, particularly in LMIC, mainly focused on gap-acceptance
behavior and critical gap estimation~\cite{cjce-2018-0796, 2461-16, PAWAR2022105582}. Gap acceptance is highly sensitive to TP~\cite{PAWAR2022105582}. Pawar et al.~\cite{PAWAR2022105582} showed that drivers under LTP and HTP accept significantly smaller gaps, resulting in elevated crash likelihoods of 127\% (LTP) and 181\% (HTP) relative to NTP. Their Generalized Linear Mixed Models (GLMM) and decision-tree (DT) analyses revealed that the TP effects on gap selection are jointly shaped by merge distance, oncoming vehicle speed, speed reduction before merging, and driver experience. These results illustrate how TP systematically modifies perceptual judgment and risk thresholds, thereby reshaping tactical decision-making at intersections.

\subsection{Time Pressure Effects on PTW Riding: Speed and Over-Speeding}

PTWs are particularly vulnerable in traffic, with over-speeding recognized as a major cause of PTW-related crashes~\cite{Sharma21042025, GUPTA2022105820, PAWAR2022105582}. Steg and van Brussel~\cite{STEG2009503} found that subjective norms and favourable attitudes toward speeding significantly increased young moped riders to over-speed, illustrating the role of cognitive and social influences.

Other recent PTW-specific studies show that TP amplifies these vulnerabilities. TP increases over-speeding tendencies due to psychological constructs such as travel-time anxiety, hurriedness, and perceived pressure to reach early~\cite{Seefong2024, Sharma21042025, GUPTA2022105820}. Latent class analyses identify three psychologically distinct rider segments minimally anxious, moderately anxious, and highly anxious, each showing different sensitivity to TP, with highly anxious riders most likely to over-speed. These findings highlight the need to integrate psychological and behavioural information when modelling TP-related riding risk.

\subsection{Time Pressure and Distraction in MTW Riders}

TP represents a cognitive state in which the perceived time available is insufficient to complete the riding task~\cite{GUPTA2022105820, PAWAR2022105582}. Under TP, riders must process information more rapidly, often with reduced cognitive reserves. This accelerated decision environment narrows attention, degrades situational awareness, and increases susceptibility to operational errors~\cite{PAWAR2020105401}. Although distraction and TP are distinct constructs, their interaction helps reveal how cognitive stress reshapes rider attention. Gupta et al.~\cite{GUPTA2022105820} found that TP can suppress intentional distractions (e.g., voluntary mobile use) due to heightened task focus, while simultaneously amplifying unintentional cognitive lapses driven by stress and mental workload. These findings reinforce that TP fundamentally alters cognitive processing, attention allocation, and behavioural stability factors central to TP prediction.

\subsection{Simulator Validity Under Time Pressure}

Driving simulators provide a safe, controlled environment for studying TP effects, eliminating real-world collision risks while maintaining experimental validity~\cite{PAWAR202229, Bham04032018}. Research consistently demonstrates that simulator behavior under TP closely mirrors real-world responses exhibiting what is termed relative validity~\cite{PAWAR202229}. These studies show that although absolute behavioral magnitudes may differ between simulated and natural environments, the fundamental patterns of TP-induced changes remain consistent: increased speeds, elevated braking intensity, and amplified vehicular variability occur in both settings. This evidence confirms that simulators are well-suited for TP research when predictive models account for systematic calibration differences between simulated and real-world data.

\subsection{Machine Learning for Behaviour and Risk Modeling}

In parallel to behavioural research, the ITS community has increasingly adopted ML and DL approaches to model driving risk. Traditional classifiers such as Random Forests (RF), Support Vector Machines (SVM), and adaptive Deep Neural Networks (DNNs) have been used to predict high-risk manoeuvres and collision likelihood, often reporting accuracies in the range of 84--90\%~\cite{Rodegast2024, Aci2025}. More advanced sequence models, including the TST, Informer architecture~\cite{zhou2021informer}, have demonstrated a strong capability to handle long time-series dependencies relevant to driving dynamics. Despite their predictive success, these models primarily estimate external outcomes (e.g., collisions, hazardous events) and remain largely insensitive to the internal cognitive precursors such as TP, that trigger risky behaviours. This highlights an important methodological gap: existing ML systems detect risk after behavioural deviations occur rather than predicting the underlying cognitive state that gives rise to those deviations.

\subsection{Research Gap and Motivation}

Across the reviewed literature, TP consistently emerges as a major cognitive stressor that destabilizes rider behaviour and increases the risk of collision. This is particularly critical for PTW riders, who face higher vulnerability than four wheeled occupants in mixed traffic environments. Despite its established importance, TP remains a significantly underexamined factor in PTW research, especially within LMICs where high usage of PTW coincides with complex and chaotic road infrastructures.

The majority of existing research on TP has focused on four-wheeled vehicles, with limited research on its effects on PTW rider behavior. Furthermore, a key limitation of this existing work is its predominant focus on analysing the behavioural outcomes of TP, rather than developing methods to model and predict the presence of TP itself. This reactive approach limits the potential for proactive safety interventions.

To address this, our research is structured into four sequential phases. First, we collected a first-of-its-kind, high-resolution PTW simulator dataset comprising 129,209 feature windows extracted via sliding window segmentation from 153 rides by 51 participants under three TP conditions (NTP, LTP, HTP). Second, we conducted an empirical analysis of TP impact to establish how it manifests in rider behavior. Third, we developed MTPS to predict TP from sensor data. Finally, we established a pathway to proactive safety systems under the Safe System Approach (SSA) using MTPS-predicted TP for collision prediction and threshold-based rider states for graded ITS interventions.

\section{Methodology for Data Collection}
Based on the gaps identified in the literature review, we designed a controlled simulator study to collect high-resolution PTW riding data under varying TP conditions. This section describes the experimental setup, participant recruitment, TP scenarios, and data collection protocol.

\subsection{Experimental Design}

\subsubsection{Simulator Setup}
\label{subsec:simulator_setup}

All experimental trials are conducted using a static PTW simulator (Technotrove Simulation Pvt. Ltd.~\cite{Technotrove}) housed at the Indian Institute of Technology Indore. The simulator is built upon an instrumented Honda motorcycle frame equipped with fully functional throttle, brake, clutch, and gear controls, thereby replicating the ergonomics of actual on-road riding. A three-screen immersive display system surrounds the rider to provide a wide field of view and realistic roadway visualization. The simulator is integrated with a real-time data acquisition system that continuously records rider inputs, vehicle dynamics, and environmental parameters using high-precision industrial-grade sensors with documented accuracy specifications. Complete technical specifications, including sensor accuracy ratings per IEC 393 standards, are provided in Table~\ref{tab:simulator_specs}. This configuration provides a high-fidelity yet safe experimental environment, allowing systematic investigation of rider responses under various traffic and roadway scenarios.

\begin{table}[htbp]
\centering
\caption{Technical Specifications of the PTW Simulator}
\label{tab:simulator_specs}
\begin{tabular}{@{}lp{0.65\linewidth}@{}}
\toprule
\textbf{Component} & \textbf{Specification} \\
\midrule

Platform & Static motorcycle frame, ISO-compliant \\
Controls & Throttle, clutch, hydraulic brakes, 5-speed gearbox, steering, handbrake \\

Sensors & Servo Potentiometers (Wire Wound – 50WW): \\
& Resistance: 5 k$\Omega$ (±10\%); Independent Linearity: ±0.5\% (IEC 393), Electrical Angle: 355° ± 3°; Mechanical Rotation: 360° \\
& Power Rating: 3 W (@ 70°C); Rotational Life: 2,000,000 revolutions \\
& Operating Temperature: –40°C to +105°C \\
& Load Cells: Brake force measurement \\
& Optical Encoders: Gear position and control tracking, Limit Switches (SPDT): Gear shift detection \\

Visual System & 3 × 50" inch LED displays, 180° field of view \\
Motion System & Electric Actuators (×4): Max acceleration ±1 G, 0–100 Hz bandwidth \\
& Stroke Length: 38.1 mm; Load Capacity: 114 kg per actuator \\

Computer Systems & Rider Station: Intel i7, 32GB RAM, GTX 1650 \\
& Instructor Station: Intel i5, 16GB RAM, GTX 1650 \\

Software & TechnoSim (AI traffic, environment modelling, scenario scripting) \\
Data Logging & Real-time data acquisition, sampling frequency $f_s \geq 100$ Hz \\

\bottomrule
\end{tabular}
\end{table}

\subsubsection{Simulator Scenario Design}
\begin{figure}[htbp]
    \centering
    \subfloat[AI vehicle path]{%
        \includegraphics[width=0.48\linewidth]{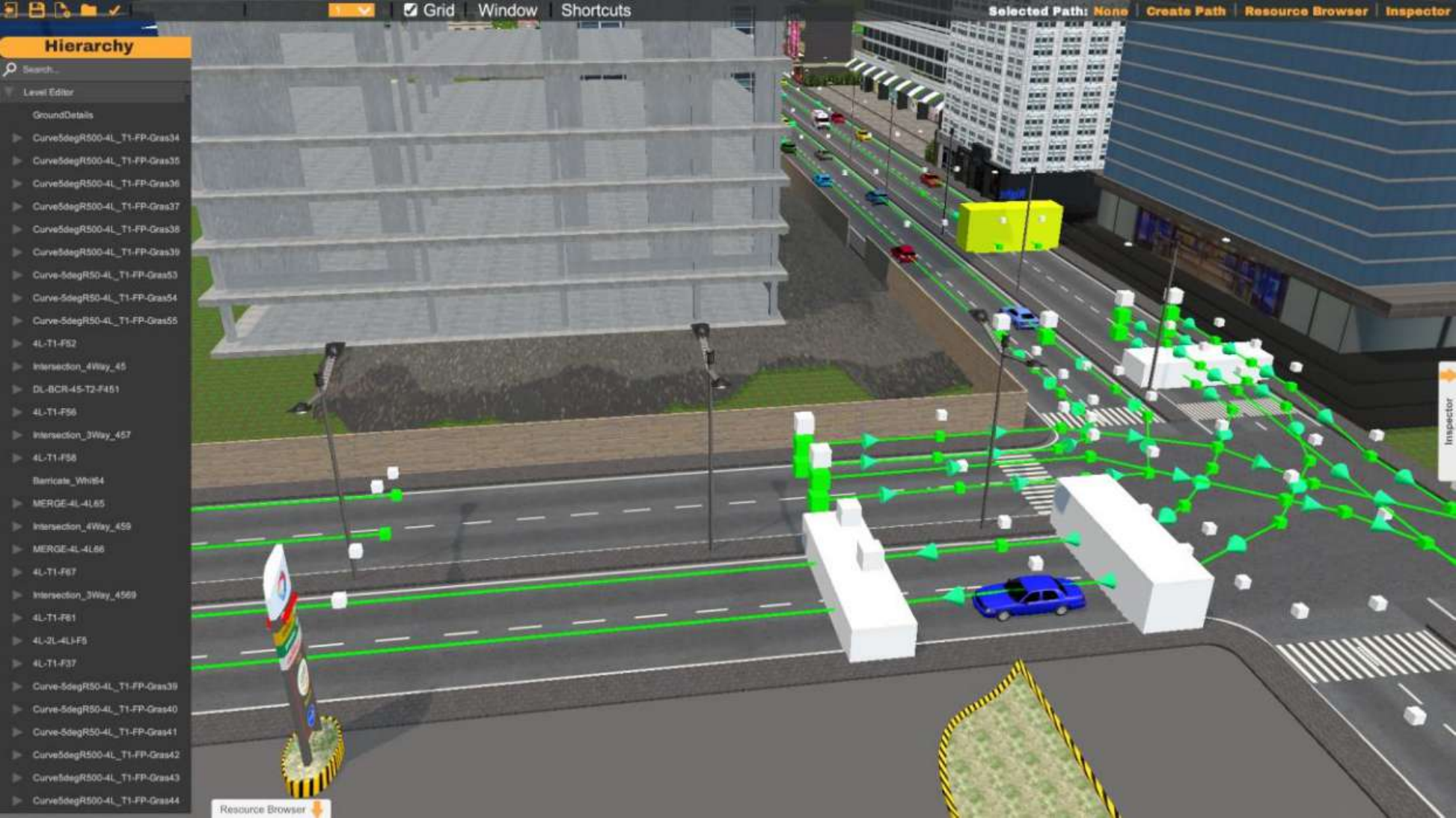}%
    }
    \hspace{2pt}
    \subfloat[Trigger events and path]{%
        \includegraphics[width=0.48\linewidth]{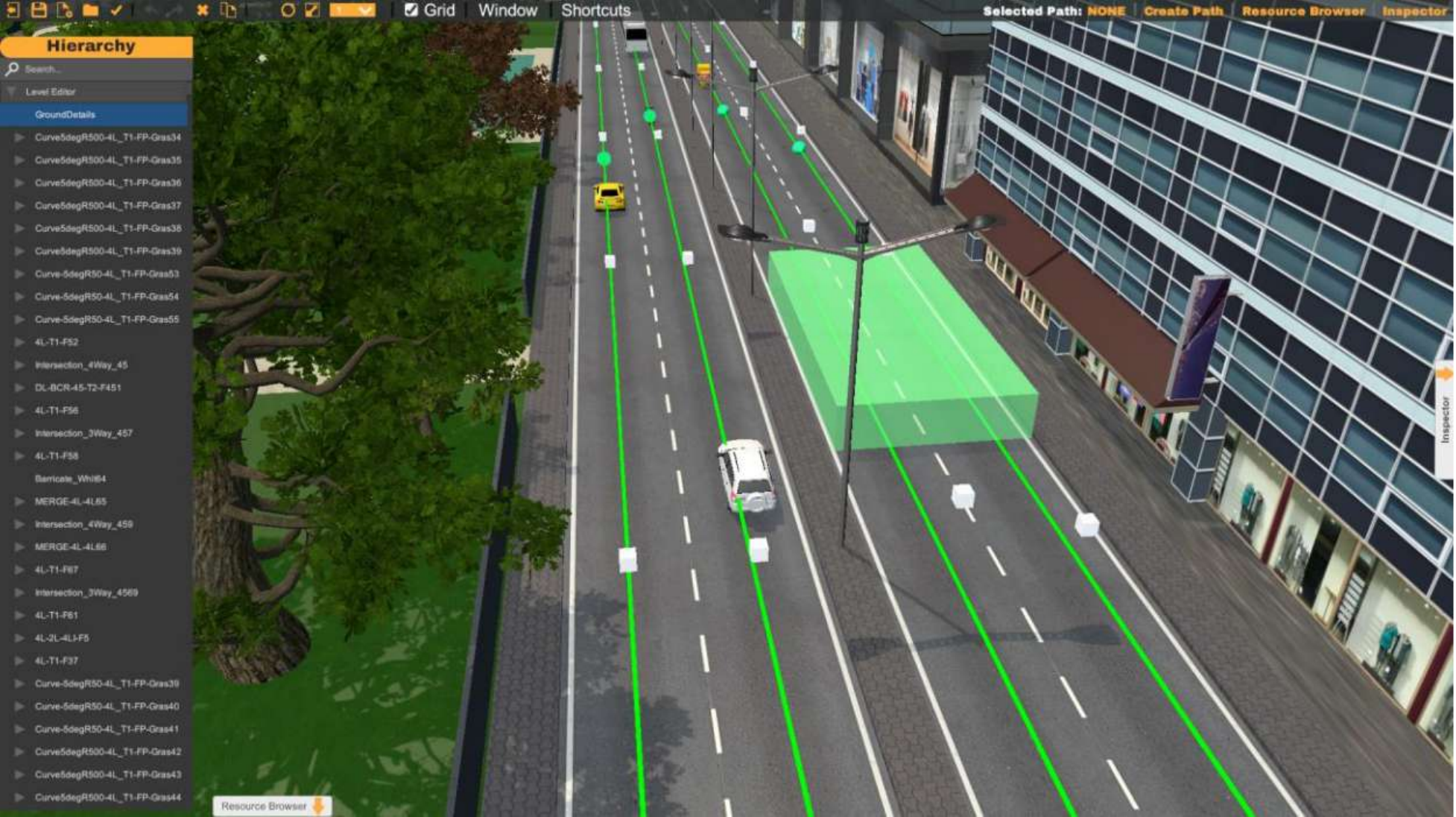}%
    }

    \vspace{2pt} % reduced from \baselineskip
    \subfloat[Intersections]{%
        \includegraphics[width=0.48\linewidth]{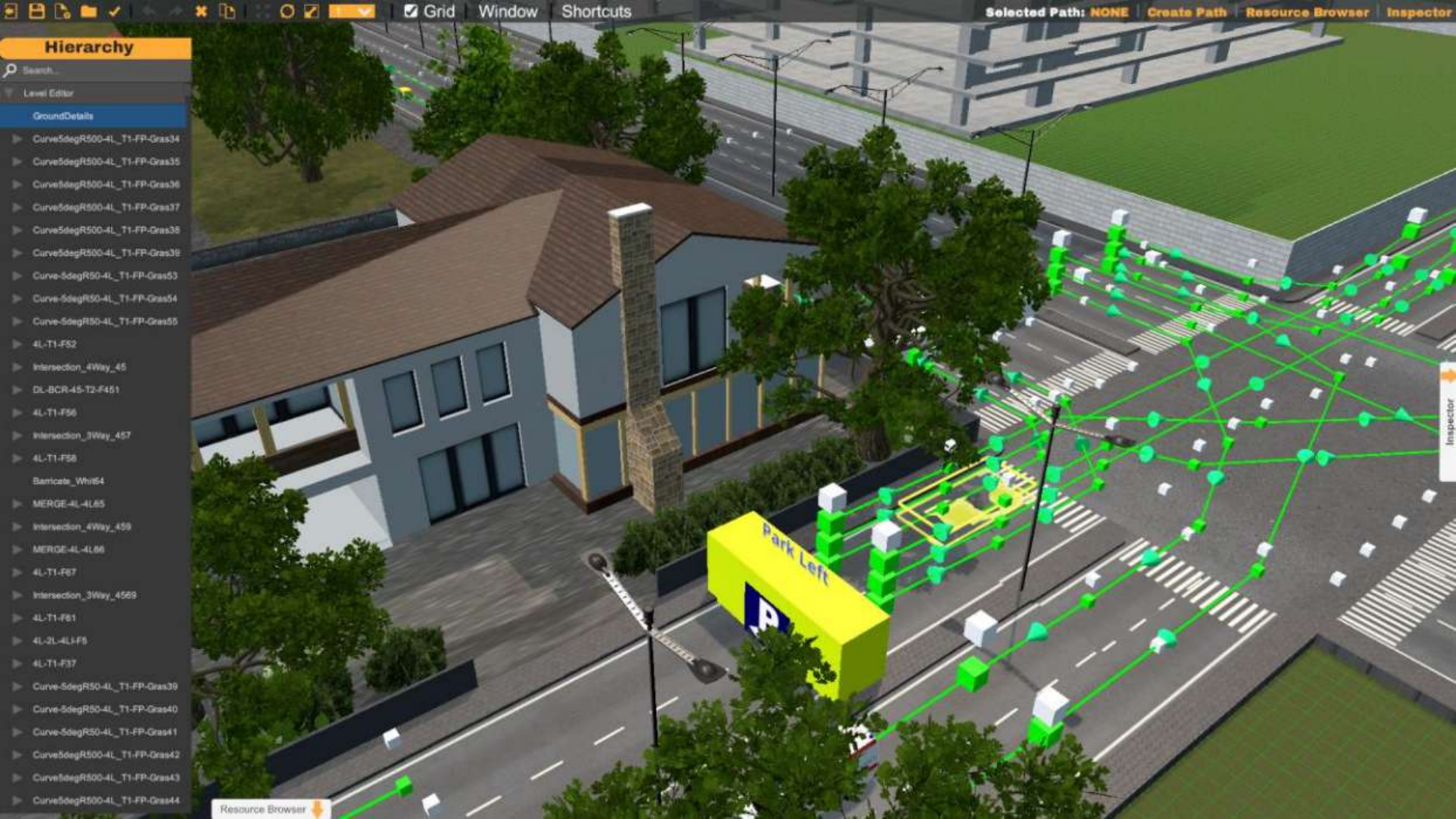}%
    }
    \hspace{2pt}
    \subfloat[Rider pathway]{%
        \includegraphics[width=0.48\linewidth]{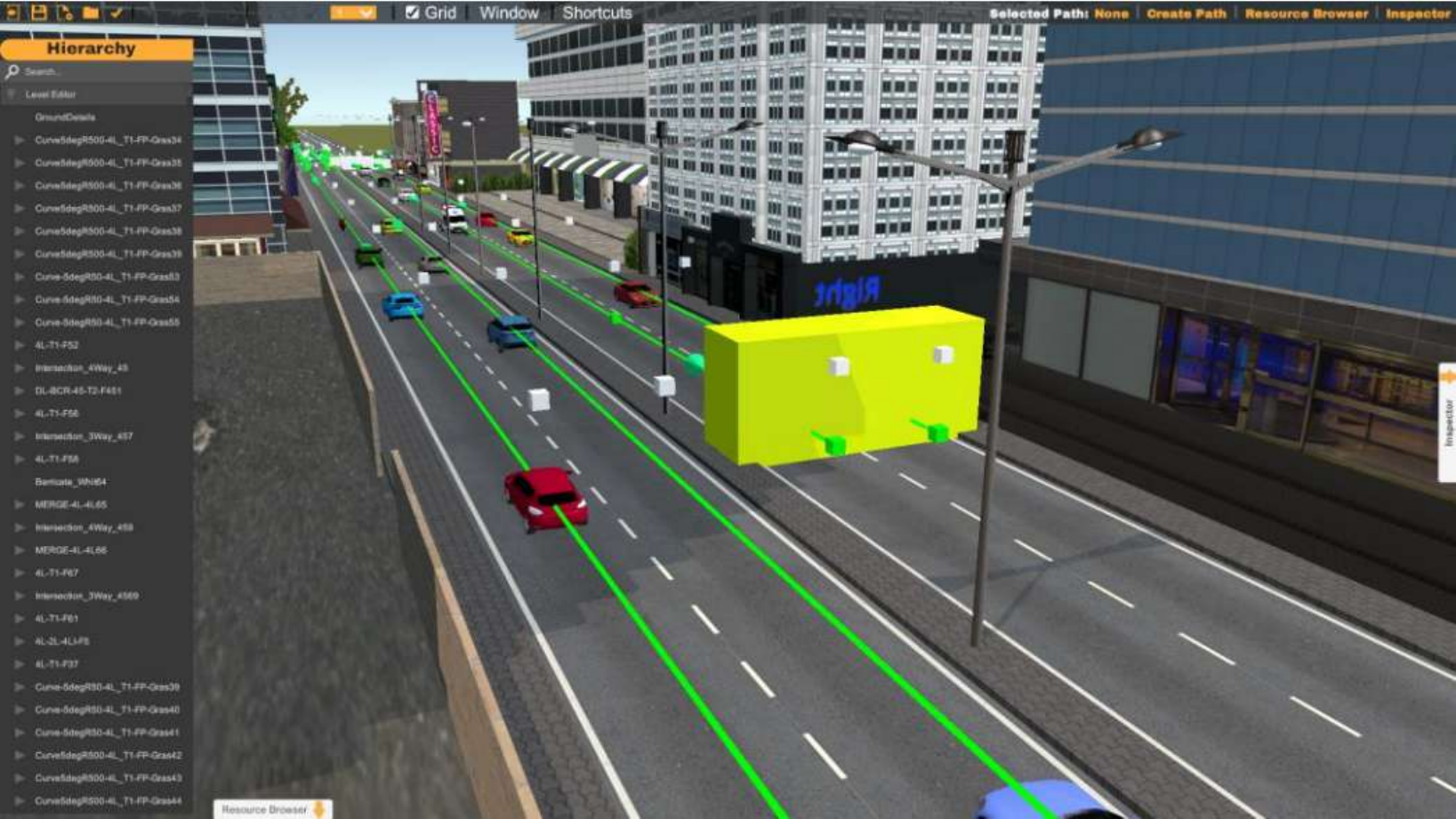}%
    }

    \caption{Sample snapshots of the simulated riding environment.}
    \label{fig:simulated_scenarios}
\end{figure}

The scenario development is carried out using the Technosim simulator platform~\cite{Technotrove}. It creates a high-fidelity urban riding environment with pedestrian crossings, intersections, obstacle overtaking, and variable traffic densities. A 4.8~km route simulates both an undivided two-lane and a four-lane divided highway with a 50~km/h speed limit, ensuring ecological validity and repeatable events that require decision-making, risk assessment, and adaptive control under varying TP levels. Fig.~\ref{fig:simulated_scenarios} shows the snapshots of scenario development (vehicle paths, trigger events, route changes, and rider trajectories).

\subsubsection{Participants}
\label{subsec:participants}

\begin{figure}[htbp]
    \centering
    % First row
    \subfloat[Participant~1]{%
        \includegraphics[width=0.46\linewidth]{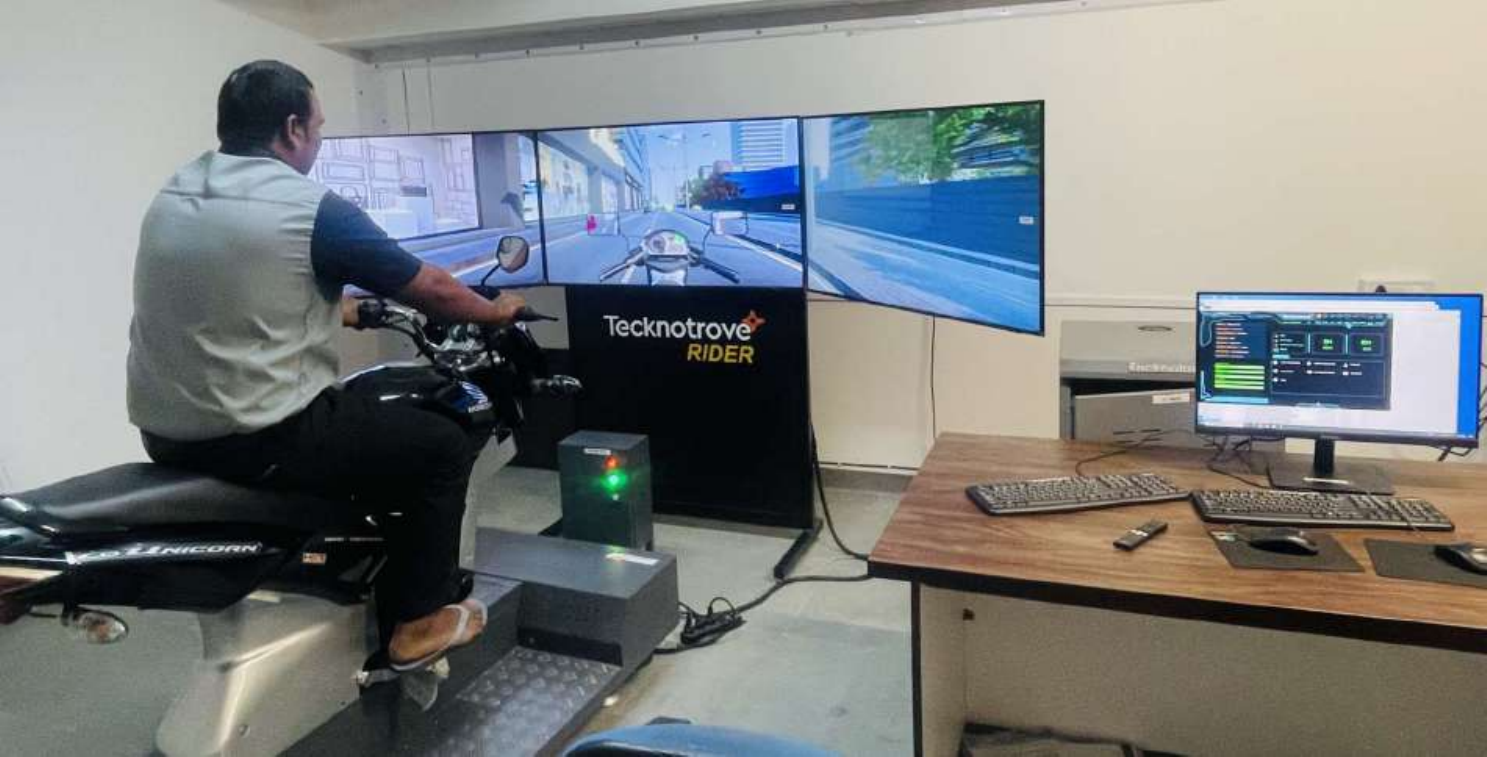}%
    }
    \hspace{1pt}
    \subfloat[Participant~2]{%
        \includegraphics[width=0.46\linewidth]{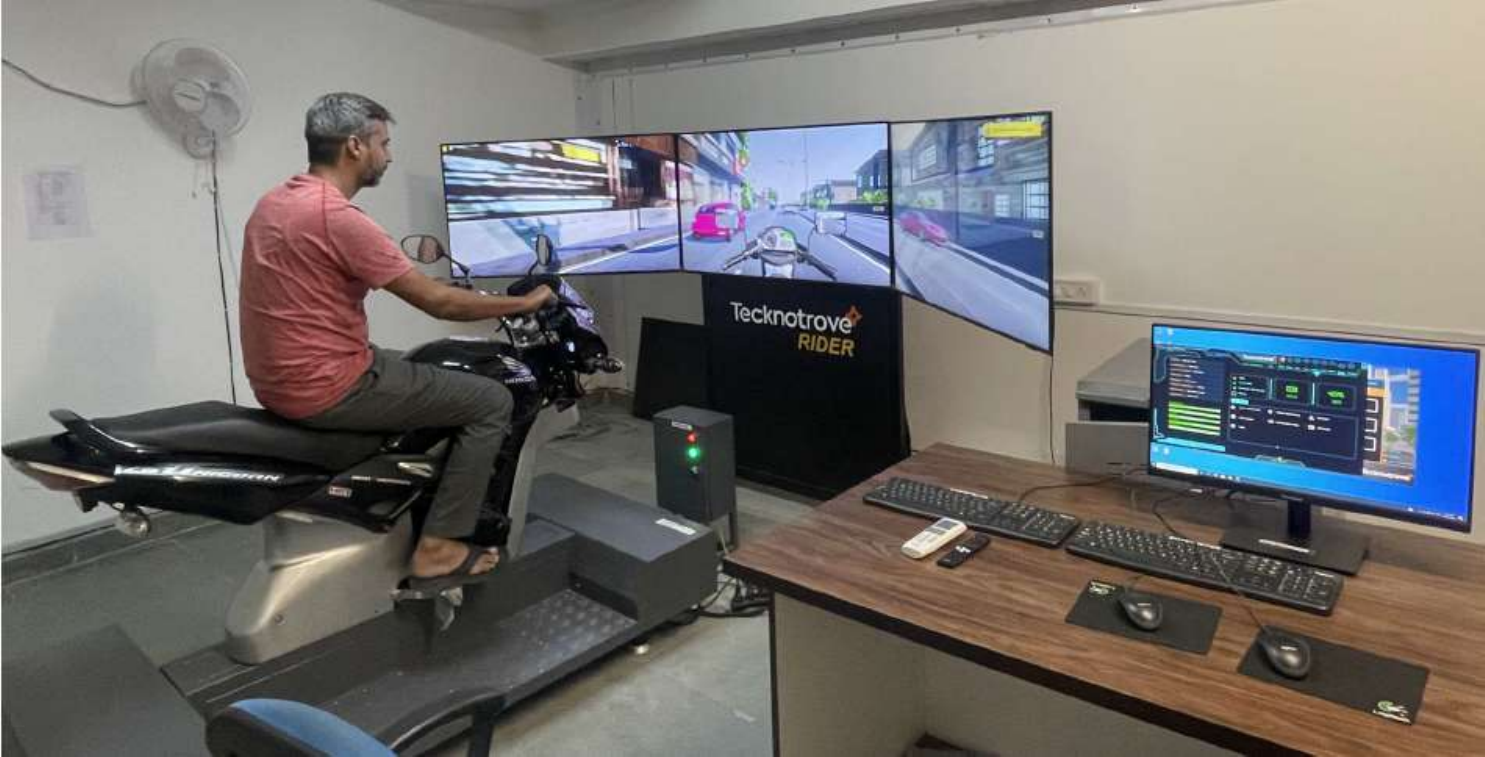}%
    }

    % Second row
    \vspace{1pt}
    \subfloat[Participant~3]{%
        \includegraphics[width=0.46\linewidth]{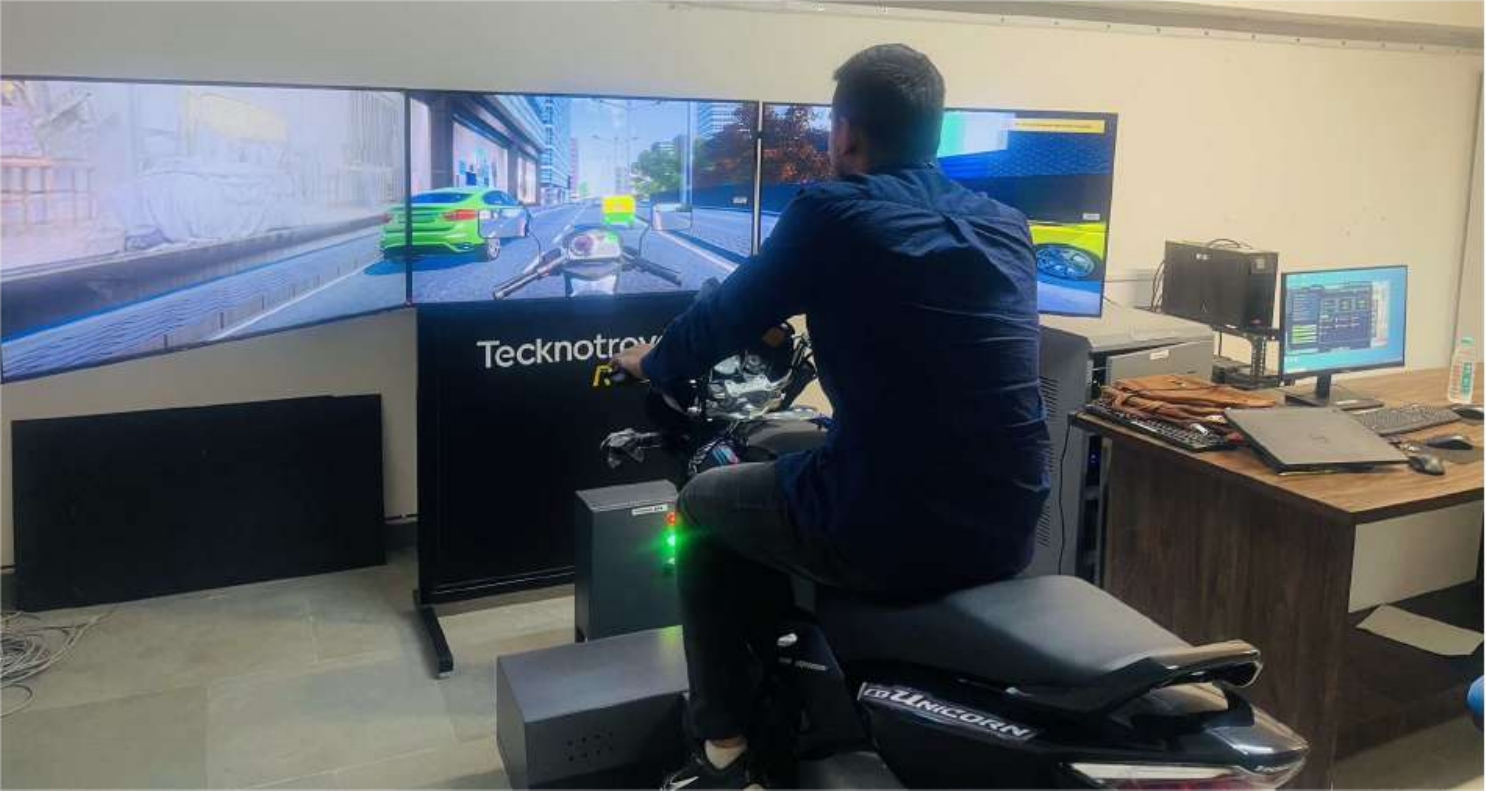}%
    }
    \hspace{1pt}
    \subfloat[Participant~4]{%
        \includegraphics[width=0.47\linewidth]{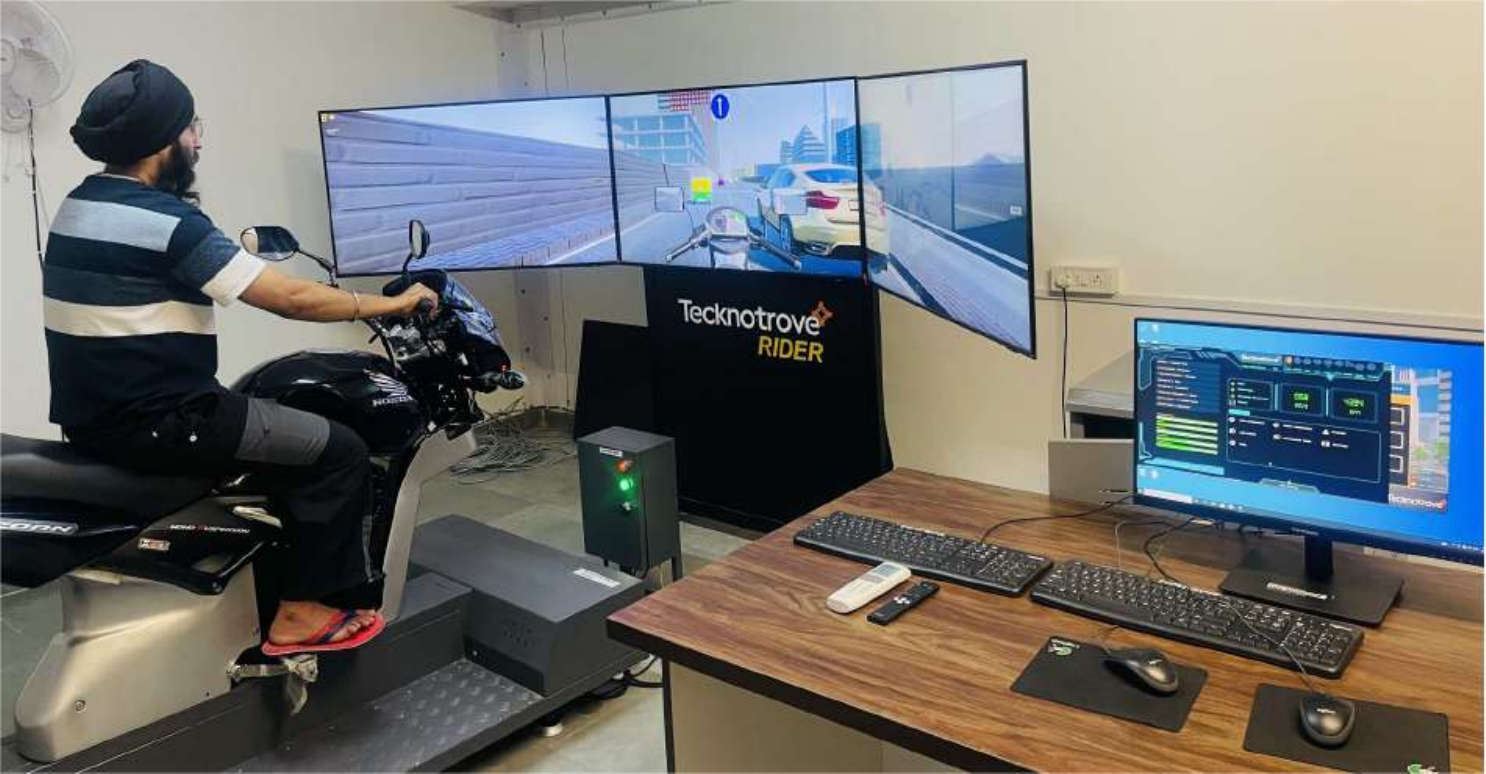}%
    }

    \caption{Sample participant snapshots from two-wheeler simulator sessions.}
    \label{fig:participants_performing_test}
\end{figure}

The vast majority of two-wheeler riders in India are male, as reflected in MoRTH road fatality statistics~\cite{morth2019,morth2020,morth2021,morth2022,morth2023}. According to official MoRTH reports, males accounted for 86.0\%, 87.3\%, 86.4\%, 86.2\%, and 85.2\% of PTW fatalities between 2019 and 2023, respectively, with most deaths in the 18--45 age group. This overwhelming male predominance provides a strong empirical basis for focusing on \textit{male participants} in this study. Accordingly, fifty-one male riders (ages 18--42) with at least two years of riding experience participated. Detailed rider characteristics are presented in Table~\ref{tab:participant_stats}, and Fig.~\ref{fig:participants_performing_test} shows participants performing simulator tasks.

\begin{table}[H]
\centering
\small
\caption{Rider Characteristics (N = 51)}
\label{tab:participant_stats}
\renewcommand{\arraystretch}{1}
\begin{tabular}{@{}lll@{}}
\toprule
Variable & Level / Description & Mean (SD) or \% \\
\midrule
Age (yrs) & 18--42 & 26.4 (5.7) \\
Experience (yrs) & $\ge 2$ & 5.6 (3.1) \\
License & Valid two-wheeler & 100\% \\
Education & Graduate / Final-year B.Tech & 58.8\% / 41.2\% \\
Simulator Exp & No / Yes & 92.2\% / 7.8\% \\
Health & Medically fit & 100\% \\
\bottomrule
\end{tabular}
\end{table}

\begin{comment}
    
\begin{table}[htbp]
\centering
\caption{Rider Characteristics (N = 51)}
\label{tab:participant_stats}
\renewcommand{\arraystretch}{1.2}
\begin{tabular}{@{}llc@{}}
\toprule
\textbf{Variable} & \textbf{Level / Description} & \textbf{Mean (SD) or \%} \\
\midrule
Age (years) & 18--42 & 26.4 (5.7) \\
Riding Experience (years) & $\ge 2$ & 5.6 (3.1) \\
License Validity & Valid two-wheeler license & 100\% \\
\midrule
Education Level & Graduate or above & 58.8\% \\
& Final-year B.Tech & 41.2\% \\
\midrule
Prior Simulator Experience & No & 92.2\% \\
& Yes & 7.8\% \\
\midrule
Health Screening & Medically fit for driving & 100\% \\
\bottomrule
\end{tabular}
\end{table}
\end{comment}

\subsubsection{TP Scenarios}
Each participant completes three simulated riding sessions under different TP conditions: NTP, LTP, and HTP. These scenarios emulate varying levels of cognitive and emotional load typical of time-sensitive situations, such as commuting to a scheduled destination. To provide a common motivational frame, participants are instructed to imagine traveling to a university examination hall.

\begin{enumerate}
    \item NTP: Ample time is provided, representing the baseline condition.  
    
    \item LTP: Riders are given 90\% of the NTP duration to complete the task (e.g., Try not to be late).  
    
    \item HTP: Riders have 80\% of baseline time, with urgency prompts (e.g., exam gate may close) to heighten stress. 

\end{enumerate}

\subsubsection{Behavioral Transition Under Exam Pressure: Why LTP Matters}

Including LTP captures the intermediate zone between safe (NTP) and risky (HTP) riding, representing early behavioral degradation. The continuum is: \textit{NTP $\rightarrow$ LTP $\rightarrow$ HTP}. NTP reflects calm, stable riding; HTP reflects high stress with overspeeding and violations. LTP acts as a threshold, capturing early risky signals such
as mild overspeeding or lane drifts, improving the model detection of rising stress, and enabling timely interventions.
 
\subsection{Data Collection}
\subsubsection{Dataset Collection and Experimental Design}

\begin{figure}[htbp]
    \centering
    \includegraphics[width=0.49\textwidth]{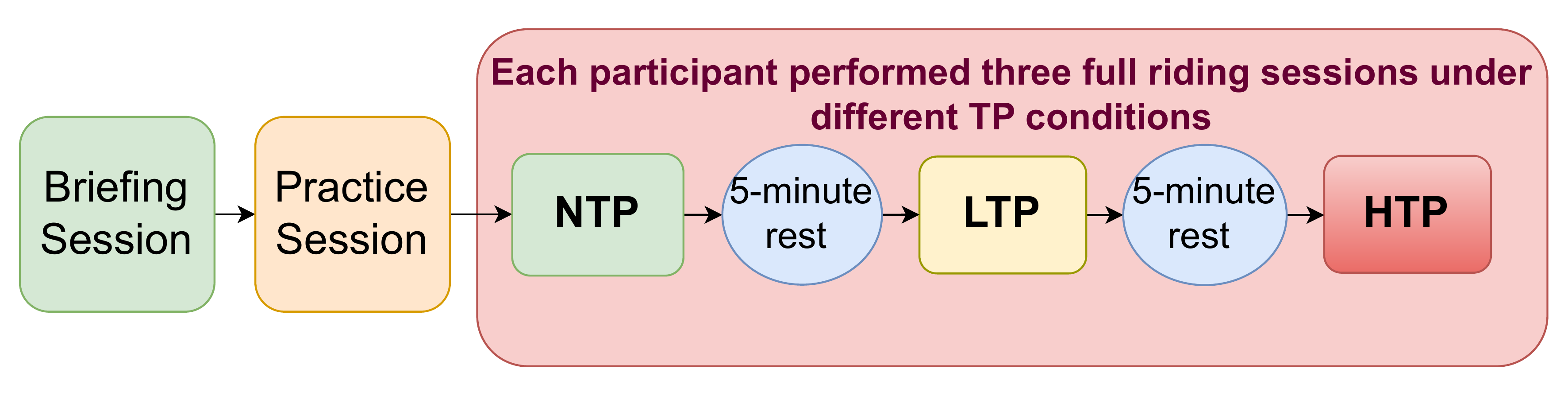}
    \caption{Flow of dataset collection and experimental design.}

    \label{fig:experiment_flowchart}
\end{figure}

Fig.~\ref{fig:experiment_flowchart} shows the experimental design and protocol with four phases: (i) Briefing: simulator familiarization and consent; (ii) Practice: 5--10~min trial ride (data excluded); (iii) Main Task: three rides under NTP, LTP, and HTP in counterbalanced order; and (iv) Rest: 5~min break to reduce fatigue. This design allows systematic assessment of rider behavior under varying TP.

\textbf{Ethics and Informed Consent:}
All participants provided informed consent prior to participation, and the study was approved by the Institutional Ethics Committee of the Indian Institute of Technology Indore (Approval No. BSBE/IITI/IHEC-11/2025/11).

\subsubsection{Dataset Details}
\label{sec:Dataset_Details}

The collected data set comprises 153 sessions from 51 participants under varying TP (NTP, LTP and HTP). Raw time-series data sampled at 100~Hz were segmented into overlapping 1-second windows (50\% overlap), yielding 129,209 feature windows across all 153 sessions. Each window contains 63 behavioral features derived from the raw sensor readings, totaling over 129,000 multivariate time-series windows. Data were collected using a high-fidelity two-wheeler simulator and include 63 features (after excluding collision-related features to prevent label leakage) (Table~\ref{tab:feature_summary}). The target variable encodes perceived TP states: $y=0$ (HTP), $y=1$ (LTP), $y=2$ (NTP). Preprocessing is performed as described in Section~\ref{subsec:dataset_preprocessing}.

\begin{table*}
\centering
\scriptsize
\caption{Summary of simulator features used in this study}
\label{tab:feature_summary}
\renewcommand{\arraystretch}{1.0} 

\small
\begin{tabular}{p{2.8cm}p{13cm}c}
\hline
\textbf{Category} & \textbf{Feature Names} & \textbf{Count} \\
\hline
\textbf{Vehicle-Controls} & Ignition, Engine, Accelerator, Brake, Clutch, Handbrake, Steering, Gear, Headlight, Horn Violation & 10 \\

\textbf{Vehicle Performance} & Speed, RPM, Fuel Economy, Distance Travelled & 4 \\

\textbf{Lighting and Indicators} & Indicator, Indicated before moving off, Indicated while turning at junction, Indicated while changing lanes, Failed to use headlights & 5 \\

\textbf{Behavioral Violations} & Over-speeding, Incorrect speed at intersections/junctions, Incorrect speed on speed breakers, Improper gap maintenance, Dangerous overtaking, Turned without indication, Incorrect lane driving, Wrong-side driving, Driving with handbrake applied, Clutch riding, Incorrect gear change sequence, Improper clutch release, Gear shift without clutch, Correct gear before moving off, Smooth releasing of clutch & 15 \\

\textbf{Traffic Rule Violations} & Crossed white line, Crossed yellow line, Crossed stop line, Signal jumping, No-entry violation, U-turn violation, No-parking violation & 7 \\

\textbf{Time Context} & Time Stamp & 1 \\

\textbf{SpatialPosition} & Position (X, Y, Z), Rotation (X, Y, Z), Lane No., Left Lane Offset, Right Lane Offset & 9 \\

\textbf{Motion and Proximity} & Lateral Velocity, Longitudinal Velocity, Headway Distance, Headway Time, Tailway Distance, Tailway Time, Leftway Distance, Rightway Distance, Steering Angle & 9 \\

\textbf{Brake Force} & Brake test done, Front Tire Brake Force, Rear Tire Brake Force & 3 \\
\hline
\end{tabular}
\\[2pt]
\footnotesize{\textit{Note: Collision-related features (e.g., collisions with vehicles, objects) were excluded from analysis to prevent label leakage.}}
\end{table*}

Collecting real-world TP data is challenging due to safety risks and ethical constraints, but we will work within these constraints. We plan to collaborate with commercial riding platforms to collect naturalistic riding data and validate MTPS in real traffic conditions as part of our future work.

\section{Methodology for TP prediction}

This section presents the architecture and mathematical formulation of MTPS.

\subsection{MTPS Architecture}
\label{subsec:mtps_architecture}

\begin{figure*}[!t]
\centering
\includegraphics[width=0.90\textwidth]{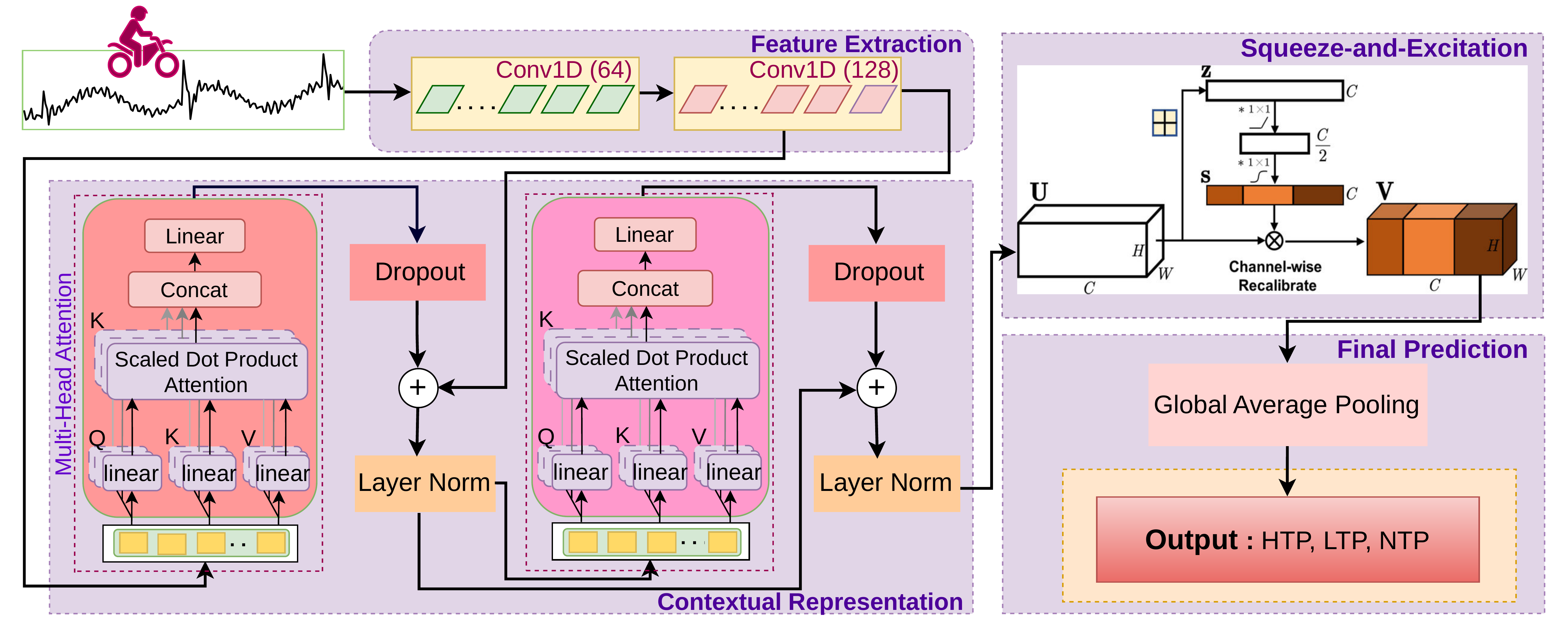}
\caption{MotoTimePressure (MTPS) architecture for predicting rider time-pressure states.}
\label{fig:MTPS_architecture}
\end{figure*}

The proposed MTPS architecture is a DL task-specific model designed to hierarchically capture rider behavior from multivariate time series data, modeling both local temporal patterns and global contextual dependencies. Fig.~\ref{fig:MTPS_architecture} shows the architecture of MTPS for predicting rider TP states. The model takes an input sequence $X \in \mathbb{R}^{T \times d}$, where $T$ is the sequence length (number of time steps) and $d$ is the number of input features. MTPS parameterizes a conditional probability distribution $P(y \mid X; \Theta)$ on the outcome classes $y \in \mathcal{Y} = \{\text{HTP}, \text{LTP}, \text{NTP}\}$, producing the final probabilistic prediction $\hat{y}$. The architecture consists of four primary components: feature extraction, context modeling, feature recalibration, and classification.

MTPS introduces the first deep learning architecture specifically designed for rider cognitive state prediction, synergistically combines temporal convolutional networks, multi-head attention (MHA), and feature recalibration to decode cognitive TP states. The architecture is specifically designed to address the multi-scale nature of TP manifestations: Conv1D layers capture micro-behaviors (sudden braking, throttle impulses), attention mechanisms identify critical temporal segments (intersection approaches, gap assessments), and Squeeze-and-Excitation (SE) blocks dynamically prioritize the most discriminative features. Unlike generic architectures, MTPS is specifically optimized for the multi-scale temporal patterns characteristic of PTW riding behavior under cognitive stress. This integrated approach represents the first architecture optimized for real-time cognitive state inference from PTW telematics.

\begin{comment} 

\paragraph{Model Algorithm}
Algorithm~\ref{alg:mtps} summarizes how the MTPS model processes input data to produce predictions.

\begin{algorithm}[htbp]
\caption{MotoTimePressure (MTPS) Algorithm}
\label{alg:mtps}
\begin{algorithmic}[1]
\REQUIRE Input sequence $\mathbf{X} \in \mathbb{R}^{T \times d}$
\ENSURE Predicted class probabilities $\hat{\mathbf{y}} \in \mathbb{R}^3$
\STATE $\mathbf{H}^{(0)} \gets \text{Reshape}(\mathbf{X})$ \COMMENT{To $(T, 1, d)$ for Conv1D}
\STATE $\mathbf{H}^{(1)} \gets \text{ReLU}(\text{Conv1D}_{64}(\mathbf{H}^{(0)}))$
\STATE $\mathbf{H}^{(2)} \gets \text{ReLU}(\text{Conv1D}_{128}(\mathbf{H}^{(1)}))$
\STATE $\mathbf{H}^{(3)} \gets \text{MHAttention}(\mathbf{H}^{(2)}, \mathbf{H}^{(2)}, \mathbf{H}^{(2)})$
\STATE $\mathbf{H}^{(3)} \gets \text{LayerNorm}(\mathbf{H}^{(2)} + \text{Dropout}(\mathbf{H}^{(3)}))$
\STATE $\mathbf{H}^{(4)} \gets \text{MHAttention}(\mathbf{H}^{(3)}, \mathbf{H}^{(3)}, \mathbf{H}^{(3)})$
\STATE $\mathbf{H}^{(4)} \gets \text{LayerNorm}(\mathbf{H}^{(3)} + \text{Dropout}(\mathbf{H}^{(4)}))$
\STATE $\mathbf{H}^{(5)} \gets \text{SEBlock}_{128}(\mathbf{H}^{(4)})$
\STATE $\mathbf{h} \gets \text{GlobalAveragePooling1D}(\mathbf{H}^{(5)})$
\STATE $\hat{\mathbf{y}} \gets \text{Softmax}(\text{Dense}(\mathbf{h}))$
\RETURN $\hat{\mathbf{y}}$
\end{algorithmic}
\end{algorithm}
\end{comment}

\subsection{Model Structure}
\label{subsec:mtps_math}

We consider the following problem: given multivariate time-series sequences $X \in \mathbb{R}^{T \times d}$ from PTW simulator data, where $T$ is the sequence length and $d$ is the number of features, MTPS learns to predict rider TP states $y \in \{\text{HTP}, \text{LTP}, \text{NTP}\}$ through hierarchical temporal modeling.

Forward Process. The architecture processes input sequences through complementary pathways that capture both local temporal patterns and global contextual dependencies. The model begins with temporal feature extraction using stacked 1D convolutional layers to capture short-term riding maneuvers:
\begin{equation}
\small
H^{(1)} = \text{ReLU}(W^{(1)} \ast X + b^{(1)}), \quad H^{(2)} = \text{ReLU}(W^{(2)} \ast H^{(1)} + b^{(2)})
\end{equation}
where $H^{(1)} \in \mathbb{R}^{64 \times T}$, $H^{(2)} \in \mathbb{R}^{128 \times T}$, $\ast$ denotes 1D convolution, and $W^{(l)} \in \mathbb{R}^{k_l \times C_{l-1} \times C_l}$. These layers extract local TP signatures such as sudden braking, rapid throttle inputs, and steering oscillations that manifest as immediate stress responses. For multi-scale global context modeling, the architecture employs two stacked multi-head attention blocks with residual connections. The first attention block processes CNN features to capture initial temporal dependencies:
\begin{equation}
\begin{array}{l}
\text{MHA}_1(H^{(2)}) = \text{Concat}(\text{head}_1^1, \dots, \text{head}_h^1) W^{O_1}; \\
Z^{(1)} = \text{LayerNorm}(H^{(2)} + \text{Dropout}(\text{MHA}_1(H^{(2)})))
\end{array}
\end{equation}

where each head $i$ computes:
\begin{equation}
\text{head}_i^1 = \text{Softmax}\left(\frac{H^{(2)}W_i^{Q_1} (H^{(2)}W_i^{K_1})^T}{\sqrt{d_k}}\right)H^{(2)}W_i^{V_1}
\end{equation}
with projection matrices $W_i^{Q_1}, W_i^{K_1} \in \mathbb{R}^{C \times d_k}$ and $W_i^{V_1} \in \mathbb{R}^{C \times d_v}$. The second attention block deepens temporal understanding by processing refined features:
\begin{equation}
\begin{array}{l}
\text{MHA}_2(Z^{(1)}) = \text{Concat}(\text{head}_1^2, \dots, \text{head}_h^2) W^{O_2}; \\
Z^{(2)} = \text{LayerNorm}(Z^{(1)} + \text{Dropout}(\text{MHA}_2(Z^{(1)})))
\end{array}
\end{equation}
where each head $i$ computes:
\begin{equation}
\text{head}_i^2 = \text{Softmax}\left(\frac{Z^{(1)}W_i^{Q_2} (Z^{(1)}W_i^{K_2})^T}{\sqrt{d_k}}\right)Z^{(1)}W_i^{V_2}
\end{equation}
with projection matrices $W_i^{Q_2}, W_i^{K_2} \in \mathbb{R}^{C \times d_k}$ and $W_i^{V_2} \in \mathbb{R}^{C \times d_v}$. This dual-attention mechanism enables the model to focus on critical TP-sensitive periods at multiple temporal scales, such as intersection approaches, lane change initiation, and traffic gap assessments, while maintaining training stability through residual connections and layer normalization.

Feature recalibration employs a Squeeze-and-Excitation block to dynamically emphasize the most discriminative features for TP detection. Given feature maps $U = Z^{(2)} \in \mathbb{R}^{T \times C}$, global average pooling squeezes temporal information:
\begin{equation}
z_c = \frac{1}{T} \sum_{t=1}^{T} u_c(t), \quad c=1,\dots,C
\end{equation}
Excitation via a gating mechanism with bottleneck:
\begin{equation}
s = \sigma(W_2 \text{ReLU}(W_1 z))
\end{equation}
where $W_1 \in \mathbb{R}^{\frac{C}{r} \times C}$, $W_2 \in \mathbb{R}^{C \times \frac{C}{r}}$, and $r$ is the reduction ratio. Features are recalibrated as $\tilde{u}_c = s_c \cdot u_c$. This mechanism amplifies TP-indicative signals like throttle variability and braking patterns while suppressing less relevant channels. The recalibrated features are aggregated through global average pooling and classified via softmax:
\begin{equation}
h = \frac{1}{T}\sum_{t=1}^{T}\tilde{u}_t \in \mathbb{R}^C; \quad \hat{y} = \text{softmax}(W_{\text{cls}}h + b_{\text{cls}})
\end{equation}
where $W_{\text{cls}} \in \mathbb{R}^{3 \times C}$ and $b_{\text{cls}} \in \mathbb{R}^{3}$. The model optimizes categorical cross-entropy to distinguish between HTP, LTP, and NTP states:
\begin{equation}
\mathcal{L} = -\frac{1}{N} \sum_{i=1}^{N} \sum_{c=0}^{2} y_{i,c} \log(\hat{y}_{i,c})
\end{equation}
where $N$ is the batch size and $y_{i,c}$ indicates ground-truth class membership.

\section{Experimental Setup}
\label{sec:experimental_setup}

\subsection{Dataset Preprocessing}
\label{subsec:dataset_preprocessing}

Data are collected using a high-fidelity two-wheeler simulator at 100~Hz, yielding 153 sessions and over 129{,}000 multivariate time-series feature windows from 51 participants (Section~\ref{sec:Dataset_Details}). The feature matrix $X \in \mathbb{R}^{T \times d}$ ($d=63$) includes kinematics, control inputs, environmental/spatial context, violations, metadata, and derived temporal features (Table~\ref{tab:feature_summary}); collision features are excluded to avoid label leakage. Preprocessing involved removing sparse columns, standardizing numeric fields (e.g., gear \texttt{N}$\to$0), encoding categorical variables, and imputing missing values via mean/mode with forward–backward fill. Features are scaled with Min–Max or Z-score normalization: $x_{\text{norm}} = (x - \mu)/\sigma$, where $\mu$ and $\sigma$ are computed from the training set. The target $y$ denotes participant-annotated TP ($y=0$ HTP, $y=1$ LTP, $y=2$ NTP).

The target $y$ denotes participant-annotated TP (0=HTP, 1=LTP, 2=NTP).

\subsection{Baseline Models and Evaluation Metrics}

MTPS is benchmarked against various ML and DL models including: decision tree (DT), recurrent neural network (RNN), convolutional neural network (CNN), simplified hyperGraph network (SHGN)~\cite{tang2024simplifying}, time series transformer (TST)~\cite{10372632} and informer~\cite{zhou2021informer}. 
Performance metrics include classification (Accuracy, F1, Precision, Recall, ROC-AUC) and regression (MAE, MSE, $R^2$).

\subsection{Implementation and Training Details}

All models are implemented in Python using PyTorch and Scikit-learn and trained on an NVIDIA T400 GPU (4~GB VRAM). The features are standardized, and data is split into training, testing, and validation sets with an 80:10:10 stratified strategy. The Adam optimizer~\cite{kingma2014adam} with sparse categorical cross-entropy loss is used, and hyperparameters are tuned on the validation set. See Table~\ref{tab:hyperparameters} for details.

\begin{table}[htbp]
\centering
\scriptsize
\caption{Training and Hyperparameter Details (MTPS)}
\label{tab:hyperparameters}
\renewcommand{\arraystretch}{0.85} % tighter row spacing
\begin{tabular}{@{}p{1.3cm} m{0.42\linewidth} l @{}}
\toprule
\textbf{Component} & \textbf{Hyperparameter} & \textbf{Value} \\
\midrule
Data Preprocessing & Train-Test-Val Split / Stratified / Standardization & 
\parbox[t]{3.3cm}{80/10/10 / Yes / Zero-Mean, Unit Var.} \\

Architecture & Conv1D Layers / Filters / Kernel & 2 / 64,128 / 3 \\
             & Activation / MHA Blocks / Heads / Key Dim & ReLU / 2 / 4 / 32 \\
             & Dropout / SE Reduction & 0.2 / 16 \\

Optimization & Optimizer / LR / Loss / Batch Size / Epochs & 
\parbox[t]{3.3cm}{Adam / $1\times10^{-3}$ / \\ Sparse CCE  64 / 50} \\

Regularization & Early Stopping / LR Reduce Patience / Factor & 5 / 3 / 0.5 \\
\bottomrule
\end{tabular}
\renewcommand{\arraystretch}{1} % reset spacing
\end{table}

\section{Results and Analysis}
\label{sec:results}

We present our results in three parts. First, we analyze how TP systematically degrades riding behavior. Second, we evaluate MTPS's classification performance against six baselines. Third, we demonstrate the practical utility of predicted TP for collision prediction and threshold-based ITS interventions.

\subsection{Empirical Analysis of TP Impact on Speed and Control}

Our statistical analysis reveals systematic behavioral degradation under TP across multiple riding domains, consistent with earlier findings by Pawar et al.~\cite{PAWAR2022105582} for four-wheelers, but extended here to PTW riders. As quantified in Table~\ref{tab:behavioral_degradation}, HTP conditions induced 48\% higher mean speeds and 36\% greater speed variability compared to NTP. All speed values are in km/h. Brake forces are recorded as normalized outputs (0–100\%) from the simulator's braking sensors, where the 604\% increase in front brake application under HTP indicates substantially more aggressive braking behavior. The progressive increase in both mean speed and standard deviation across NTP $\rightarrow$ LTP $\rightarrow$ HTP conditions demonstrates that TP systematically degrades riding control and increases risk exposure. Control patterns showed equally dramatic degradation, with 604\% higher front brake force and 35\% increased gear usage under HTP conditions.

\begin{table}[h!]
\centering
\caption{Systematic Behavioral Degradation Under TP}
\label{tab:behavioral_degradation}
\footnotesize
\begin{tabular}{lcccc}
\toprule
\textbf{Metric} & \textbf{NTP} & \textbf{LTP} & \textbf{HTP} & \textbf{NTP → HTP } \\
\midrule
\textbf{Speed Characteristics} & & & & \\
\quad Mean Speed (km/h) & 33.12 & 39.69 & 49.00 & +48.0\% \\
\quad Speed Variability, SD & 19.42 & 21.67 & 26.48 & +36.4\% \\
\midrule
\textbf{Control Patterns} & & & & \\
\quad Mean Gear Usage & 2.78 & 3.22 & 3.75 & +35.0\% \\
\quad Front Brake Force & 0.021 & 0.046 & 0.148 & +604\% \\
\quad Rear Brake Force  & 1.91 & 2.29 & 2.87 & +50\% \\
\bottomrule
\end{tabular}
\end{table}

\subsection{The Inadequacy of Rider Compensation Under TP}

TP significantly compromises riding stability. As shown in Table~\ref{tab:tp_danger}, riders under HTP have 21.3\% worse control, yet commit 10.2\% more dangerous mistakes compared to the NTP condition. This indicates that even when riders attempt to compensate for degraded control, they cannot fully prevent unsafe events. In this analysis, control variability is measured as the standard deviation of the steering wheel angle (in degrees), and safety violations represent the total count of critical errors such as dangerous overtaking, signal violations, improper gap maintenance, and lane-crossing errors per ride. Values are presented as mean ± standard deviation. These metrics directly quantify how degraded control translates into behaviours that increase crash risk.

The data reveal a clear pattern: TP $\rightarrow$ degraded control $\rightarrow$ attempted self-correction $\rightarrow$ persistent increase in serious violations. This natural compensatory response has three critical limitations: it is reactive (initiated only after the control deteriorates), incomplete (unable to eliminate all hazardous behaviours), and unstable (collapses under high pressure). Consequently, TP creates measurable collision risks that inherent rider adaptation cannot sufficiently mitigate.

\begin{table}[h!]
\centering
\caption{TP Degrades Control.} 
\label{tab:tp_danger}
\begin{tabular}{lccc}
\hline
\textbf{Metric} & \textbf{HTP} & \textbf{NTP} & \textbf{Change} \\
\hline
Control Variability & 9.08 ± 8.00 & 7.48 ± 6.80 & +21.3\% \\
Safety Violations & 6.31 ± 6.12 & 5.73 ± 5.57 & +10.2\%\\
\hline
\end{tabular}
\end{table}

\subsection{Systemic Risk Escalation Under TP}

TP triggers dangerous behavioral changes across multiple riding domains, creating compound collision risks. As Table~\ref{tab:tp_comprehensive} demonstrates, HTP conditions cause severe performance degradation: riders execute 58\% more turns at incorrect speeds at intersections, exhibit 36\% more sudden braking incidents, apply 50\% higher rear brake forces, and show 67\% less frequent clutch riding compared to NTP. 

This multi-system failure pattern reveals that TP doesn't merely affect isolated skills but compromises the entire riding ecosystem. The combination of rushed decision-making (risky turns), impaired emergency responses (sudden braking), and degraded vehicle control (reduced clutch engagement) creates particularly hazardous conditions where multiple safety systems fail simultaneously. These interconnected risk factors dramatically increase collision risk.

\begin{table}[h!]
\centering
\caption{TP-Induced Behavioral and Control Risks}
\label{tab:tp_comprehensive}
\begin{tabular}{lccc}
\hline
\textbf{Risk Category} & \textbf{NTP} & \textbf{HTP} & {Change} \\
\hline
Risky Turns (Incorrect-Speed) & 0.73 & 1.15 & +58\% \\
No-Signal Turns at Intersections & 1.76 & 2.11 & +20\% \\
Sudden Braking Incidents & 1.32 & 1.80 & +36\% \\
Rear Tire Brake Force & 1.91 & 2.87 & +50\% \\
Clutch Riding Frequency & 0.43 & 0.14 & -67\% \\
\hline
\end{tabular}
\end{table}

Risky turns and no-signal turns at intersections represent the average number of high-risk events per ride. No-signal turns at intersections represent another key procedural violation, increasing by 20\% under HTP. Sudden braking represents the average count of incidents. Rear Tire Brake Force denotes the mean brake force measured in sensor units. Clutch Riding Frequency reflects instances of keeping the clutch partially engaged while riding; the 67\% decrease under HTP may indicate abandonment of proper clutch technique rather than improved control.

Consistent with the literature and confirmed by our empirical analysis, TP significantly increases collision risk and control instability, as evidenced by the consistent deviations in speed control, brake application, and clutch management. These findings provide a clear empirical basis for the MTPS framework, demonstrating the need to predict cognitive stress (TP) to enable proactive intervention.

\subsection{MTPS Model Performance and Validation}

\subsubsection{Overall Performance Comparison}

\begin{table*}[t]
\caption{Results comparison of all proposed models including MTPS. We evaluate each model five times and report the mean (standard deviation).}
\label{tab:performance_comparison}
\centering
\footnotesize
\setlength{\tabcolsep}{4pt} % reduce column spacing for readability
\begin{tabular}{lcccccccc}
\toprule
\textbf{Model} & \textbf{Accuracy (\%)} & \textbf{F1 (\%)} & \textbf{Prec. (\%)} & \textbf{Rec. (\%)} & \textbf{ROC (\%)} & \textbf{MAE} & \textbf{MSE} & \textbf{R²} \\
\midrule
DT & 79.21 ± 0.58 & 79.50 & 81.68 & 79.21 & 93.83 & 0.2291 & 0.2716 & 0.6070 \\

RNN & 81.97 ± 0.49 & 82.05 & 84.25 & 81.97 & 95.37 & 0.2194 & 0.2975 & 0.5695 \\

CNN & 85.61 ± 0.54 & 85.52 & 85.95 & 85.61 & 97.21 & 0.1678 & 0.2155 & 0.6881 \\

SHGN & 89.33 ± 0.45 & 89.41 & 90.26 & 89.33 & 98.27 & 0.1107 & 0.1188 & 0.8281 \\

TST & 89.40 ± 0.28 & 89.45 & 90.42 & 89.40 & 98.35 & 0.1138 & 0.1294 & 0.8127 \\

Informer & 90.11 ± 0.39 & 90.11 & 90.10 & 90.11 & 98.06 & 0.1105 & 0.1336 & 0.8066 \\

\textbf{MTPS (Proposed)} & \textbf{91.53 ± 0.27} & \textbf{91.54} & \textbf{91.56} & \textbf{91.53} & \textbf{98.93} & \textbf{0.0928} & \textbf{0.1090} & \textbf{0.8423} \\
\bottomrule
\end{tabular}
\end{table*}

The proposed MTPS model is benchmarked against six diverse baseline models spanning traditional machine learning, deep learning, and state-of-the-art architectures (SHGN, TST, Informer). As shown in Table~\ref{tab:performance_comparison}, MTPS achieves SOTA performance across all metrics with 91.53\% accuracy and 98.93\% ROC AUC, alongside the lowest MAE (0.0928) and highest $R^{2}$ (0.8423). This consistent improved performance across fundamentally different architectural paradigms demonstrates the architectural robustness of MTPS beyond any single modeling approach.

\subsubsection{Ablation Study}
\label{subsec:ablation_study}

To validate the architectural design of MTPS, we conducted a comprehensive ablation study by systematically removing key components. The results presented in Table~\ref{tab:ablation_study} demonstrate that each element contributes significantly, with the full integrated architecture achieving optimal results. Removing attention causes the largest drop (0.81\%), followed by SE blocks (0.64\%), while Conv-only reduces accuracy by 1.11\%.

\begin{table}[htbp]
\centering
\caption{Ablation study evaluating the contribution of MTPS components.}
\label{tab:ablation_study}
\begin{tabular}{lcc}
\toprule
\textbf{Model Variant} & \textbf{Accuracy (\%)} & \textbf{$\Delta$ (\%)} \\
\midrule
Full MTPS  & 91.53 & \textbf{--} \\

w/o Squeeze-Excitation & 90.89 & -0.64 \\
w/o Attention & 90.72 & -0.81 \\
w/o Residual \& Normalization & 91.10 & -0.43 \\
Conv-only & 90.42 & -1.11 \\
\bottomrule
\end{tabular}
\end{table}

\subsubsection{Statistical Significance of MTPS}
\label{subsec:stats_mtps}

To assess significance, paired t-tests~\cite{HEDBERG2015277} and one-way ANOVA~\cite{Zhang2012TrafficSafety} are conducted on five-fold accuracy scores ($\alpha=0.05$). Results in Table~\ref{tab:mtps_stat_significance} indicate that MTPS delivers statistically significant gains over all baselines. Paired t-tests between MTPS and each baseline (DT, RNN, CNN, SHGN, TST, Informer) yield p-values $< 0.001$ and t-statistics from $14.47$ (vs. Informer) to $61.60$ (vs. DT). The one-way ANOVA comparing all models collectively produces an F-statistic of $3812.78$ with $p < 0.000001$, confirming that the observed differences in model performance are statistically significant overall. Additionally, Cohen's d effect sizes range from $9.52$ to $73.62$, indicating very large practical significance. These results robustly confirm the statistical significance of MTPS for TP classification.

\begin{table}[htbp]
\caption{Comparative analysis of MTPS model performance ($\alpha = 0.05$)}
\label{tab:mtps_stat_significance}
\centering
\begin{tabular}{@{}lccc@{}}
\toprule
\textbf{Compared Model} & \textbf{t-statistic} & \textbf{p-value} & \textbf{Significance} \\ 
\midrule
DT        & 61.60   & 0.000000   & Yes \\
RNN       & 58.39  & 0.000000   & Yes \\
CNN       & 55.72   & 0.000000   & Yes \\
SHGN      & 31.76   & 0.000006   & Yes \\
TST       & 31.27   & 0.000006   & Yes \\
Informer  & 14.47   & 0.000133   & Yes \\
\midrule
One-way ANOVA & 3812.78 & 0.000000 & Yes \\
\bottomrule
\end{tabular}
\end{table}

\begin{comment}
    
\subsubsection{Importance of TP Prediction}

The utility of MTPS-predicted TP is demonstrated by its integration into a downstream Informer crash risk model, detailed in Section~\ref{subsec:impact_tp_app}. Using predicted TP as a feature increases collision prediction accuracy from 91.25\% to 93.51\%, approaching the performance of an oracle model with ground-truth TP (93.72\%). This shows that predicted TP, which captures the cognitive why behind risky maneuvers, provides a unique signal for proactive safety systems that kinematics alone cannot offer, underscoring its critical role in rider safety management.
\end{comment}

\subsubsection{Impact of Feature Subsets on MTPS Performance}
\label{subsec:feature_subset}

To evaluate the impact of input features on MTPS performance, experiments were conducted using top-ranked subsets based on feature importance, including critical indicators such as Over-speeding, Sudden braking, Incorrect speed while turning on intersections and junctions and Dangerous overtaking. As shown in Table~\ref{tab:feature_subset_results}, accuracy progressively decreases with fewer features, dropping from 87.13\% with 21 features to 77.54\% with only 5 features. A minimal viable set of 7 features maintains 81.05\% accuracy.

\begin{table}[htbp]
\centering
\caption{MTPS performance with top-ranked feature subsets (\%)}
\label{tab:feature_subset_results}
\begin{tabular}{c c c c c c}
\toprule
\textbf{\ Features} & \textbf{Accuracy} & \textbf{F1 Score} & \textbf{Precision} & \textbf{Recall} & \textbf{ROC AUC} \\
\midrule
5  & 77.54 & 77.68 & 77.99 & 77.54 & 93.83 \\
7  & 81.05 & 81.15 & 81.31 & 81.05 & 95.46 \\
9  & 83.51 & 83.50 & 83.50 & 83.51 & 96.52 \\
21 & 87.13 & 87.20 & 87.30 & 87.13 & 97.81 \\
63 & 91.53 & 91.54 & 91.56 & 91.53 & 98.93 \\
\bottomrule
\end{tabular}
\end{table}

\subsection{Model Performance and Reliability}

\subsubsection{Confusion Matrix}

\begin{figure}
\centering
\includegraphics[width=0.50\linewidth, height=4cm]{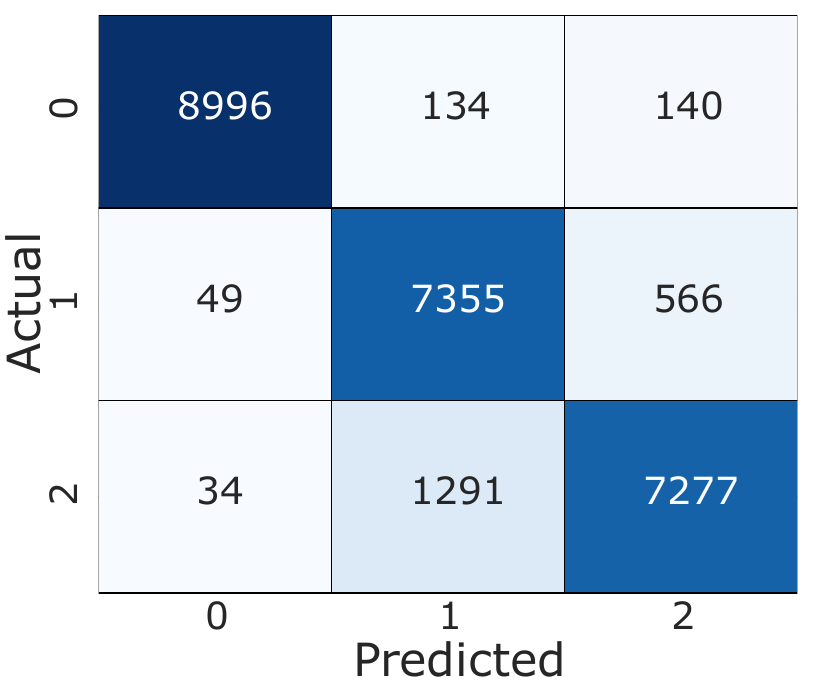}
\caption{Confusion matrix of MTPS}
\label{fig:confusion_matrix}
\end{figure}

Fig.~\ref{fig:confusion_matrix} presents the confusion matrix for MTPS. The model achieves high true positive rates across all classes, particularly for HTP (Class 0) and NTP (Class 2), indicating effective classification of distinct rider TP states.

\subsubsection{ROC and Discriminatory Power}

\begin{comment}
   
\begin{figure}[htbp]
\centering
\includegraphics[width=0.6\linewidth]{Img/MTPS_roc_curve_multiclass.pdf}
\caption{Multiclass ROC curves for MTPS.}
\label{fig:roc_curves}
\end{figure}
 \end{comment}

\begin{figure}[htbp]
  \centering
  \subfloat[Multiclass ROC.\label{fig:roc_curves}]{
    \includegraphics[width=0.45\columnwidth]{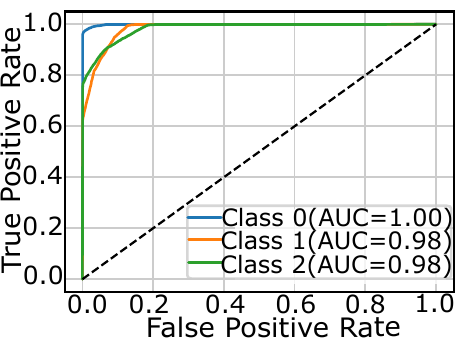}
  }
  \hfill
  \subfloat[Calibration curves.\label{fig:calibration_curves}]{
    \includegraphics[width=0.45\columnwidth]{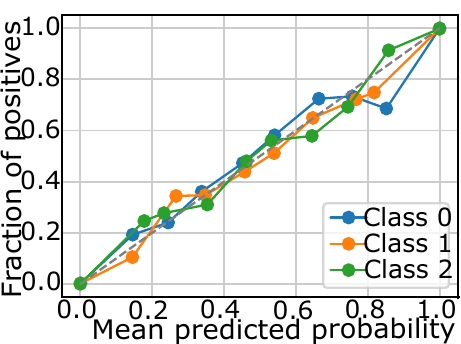}
  }
  \caption{Performance comparison: (a) ROC curves and (b) calibration curves.}
  \label{fig:roc_calibration}
\end{figure}

MTPS achieves a ROC-AUC of 98.93\%, indicating effective class discrimination. The per-class analysis (Fig.~\ref{fig:roc_curves}) reports a perfect AUC of 1.00 for HTP and excellent AUCs of 0.98 for LTP and NTP, reflecting robust and consistent performance across all classes.

\begin{table*}[h!]
\centering

\caption{Rider state transitions and statistical validation using MTPS precision and recall (95\% Wilson CIs).}

\renewcommand{\arraystretch}{1.1}
\begin{tabular}{p{1.7cm}p{4.3cm}p{4.5cm}p{6.0cm}}
\hline
\textbf{Phase / State} & \textbf{Threshold Conditions $(P)$} & \textbf{Statistical Support (Precision / Recall with 95\% CI)} & \textbf{Interpretation and ITS applications} \\
\hline
1: Calm (NTP) & $P(\text{NTP}) \geq 0.80$, $P(\text{LTP}) \leq 0.15$, $P(\text{HTP}) \leq 0.10$ 
& NTP Precision = 0.912 (0.905--0.918), Recall = 0.846 (0.838--0.853) 
& High NTP precision ensures stable and safe state detection; minimal false alarms. \\

2: Transition & $0.50 \leq P(\text{NTP}) < 0.70$, $0.25 \leq P(\text{LTP}) < 0.40$, $P(\text{HTP}) < 0.10$ 
& LTP Recall = 0.923 (0.917--0.928) 
& High LTP recall reflects uncertainty; appropriate for monitoring and watchlist. \\

3: Manageable Stress (LTP) & $P(\text{LTP}) \geq 0.65$, $P(\text{NTP}) \leq 0.30$, $P(\text{HTP}) \leq 0.10$ 
& LTP Precision = 0.837 (0.830--0.845), Recall = 0.923 (0.917--0.928) 
& Strong recall ensures reliable stress detection; lower precision favors mild, non-intrusive intervention. \\

4: Elevated Risk & $P(\text{HTP}) \geq 0.30$, $0.40 \leq P(\text{LTP}) < 0.60$, $P(\text{NTP}) \leq 0.15$ 
& HTP Recall = 0.970 (0.967--0.974) 
& High recall captures most risky riders; ideal for early-warning alerts. \\

5: Critical (HTP) & $P(\text{HTP}) \geq 0.70$, $P(\text{NTP}) \leq 0.10$ 
& HTP Precision = 0.991 (0.989--0.993), Recall = 0.970 (0.967--0.974) 
& Very high HTP precision ensures critical interventions (alarms, haptic feedback) are only triggered under reliable detection, minimizing false alarms. \\

6: Recovery (LTP) & $P(\text{LTP}) \geq 0.65$, $0.10 < P(\text{HTP}) \leq 0.20$, $0.10 < P(\text{NTP}) \leq 0.20$ 
& LTP Precision = 0.837 (0.830--0.845), Recall = 0.923 (0.917--0.928) 
& High LTP recall with moderate HTP/NTP confirms de-escalation; intervention can be relaxed. \\
\hline
\end{tabular}
\label{tab:threshold_support}
\end{table*}

\subsubsection{Probability Outputs from Softmax Classification}
MTPS is a three-class classifier representing NTP, LTP, and HTP. 
For a multivariate input $\mathbf{x}_t$ at time $t$, the model outputs raw logits $z_k(t)$ for each class $k \in \{\text{NTP},\text{LTP},\text{HTP}\}$. 
The softmax function gives class probabilities:
\begin{equation}
P(k \mid \mathbf{x}_t) =
\frac{\exp(z_k(t))}{\sum_{j \in \{\text{NTP},\text{LTP},\text{HTP}\}} \exp(z_j(t))},
\label{eq:softmax}
\end{equation}
where $P(k \mid \mathbf{x}_t)\in[0,1]$ and $\sum_k P(k \mid \mathbf{x}_t)=1$. 
Thus, MTPS outputs $\mathbf{P}(t) = [P(\text{NTP}\mid \mathbf{x}_t), P(\text{LTP}\mid \mathbf{x}_t), P(\text{HTP}\mid \mathbf{x}_t)]$. 
The predicted class and confidence are:
\begin{equation}
\hat{c}(t) = \arg\max_k P(k\mid \mathbf{x}_t), \qquad
s(t) = \max_k P(k\mid \mathbf{x}_t),
\label{eq:class_conf}
\end{equation}
where $\hat{c}(t)$ denotes the most probable TP state and $s(t)$ its confidence.

% \begin{figure}[htbp]
% \centering
% \includegraphics[width=0.65\linewidth]{Img/MTPS_roc_curve_multiclass.pdf}
%    \caption{Multiclass ROC.}
% \label{fig:roc_curves}
% \end{figure}

% \begin{figure}[htbp]
% \centering
% \includegraphics[width=0.65\linewidth]{Img/MTPS_calibration_curve_multiclass.pdf}
% \caption{Calibration curves.}
%  \label{fig:calibration_curves}
% \end{figure}

\subsubsection{Calibration and Transition Across TP States}

\begin{comment}
    
\begin{figure}[htbp] 
    \centering 
    \includegraphics[width=0.6\linewidth]{Img/MTPS_calibration_curve_multiclass.pdf} 
    \caption{Calibration curves for MTPS across all three TP classes.} 
    \label{fig:calibration_curves} 
\end{figure} 
\end{comment}

The MTPS calibration curve (Fig.~\ref{fig:calibration_curves}) shows predicted probabilities $P(k \mid \mathbf{x}_t)$ closely match observed frequencies  $P(k)$, enabling their use as confidence scores for ITS interventions. Thresholds map outputs along NTP $\rightarrow$ LTP $\rightarrow$ HTP, with rider states inferred from multivariate patterns (speed, spacing, maneuvering, lane stability) to capture rising cognitive load and inform timely intervention.

\subsubsection{Model Performance and Confidence Intervals}

Class-wise precision, recall, and F1 score are computed directly from the confusion matrix (Fig.~\ref{fig:confusion_matrix}). 95\% confidence intervals (CIs) are computed using the Wilson score interval~\cite{rajaraman2022deep,binomialciwilsonwikipedia}:

\begin{table}[h!]
\centering
\caption{Class-wise MTPS performance with 95\% Wilson CIs.}
\begin{tabular}{c|ccc}
\toprule
Class & Precision (CI) & Recall (CI) & F1 \\
\midrule
HTP & 0.991 (0.989--0.993) & 0.970 (0.967--0.974) & 0.980 \\
LTP & 0.837 (0.830--0.845) & 0.923 (0.917--0.928) & 0.878 \\
NTP & 0.912 (0.905--0.918) & 0.846 (0.838--0.853) & 0.878 \\
\bottomrule
\end{tabular}
\label{tab:class_perf}
\end{table}

\subsubsection{Worked Example: Calm (NTP) Class}

We obtain calculations that support the Calm (NTP) threshold (Table~\ref{tab:threshold_support} and Fig.~\ref{fig:MTPSPREDTPFremework}) assignment ($P(\text{NTP}) \geq 0.80$, $P(\text{LTP}) \leq 0.15$, $P(\text{HTP}) \leq 0.10$). The precision, recall and confidence intervals are derived from the confusion matrix (Fig.~\ref{fig:confusion_matrix}) using 95\% CI. The confusion matrix yields $TP = 7277$, $FP = 706$, $FN = 1325$, giving precision $= 0.912$, recall $= 0.846$, and F1 $= 0.878$. The class-wise performance of the MTPS with the Wilson CIs 95\% is shown in Table~\ref{tab:class_perf}. This ensures stable baseline detection while minimizing false alarms, providing a statistically justified basis for Table~\ref{tab:threshold_support}.

\subsection{Threshold-Based Rider State Transitions and Sensitivity Analysis}

\subsubsection{Threshold Selection and Statistical Support}
The threshold values in Table~\ref{tab:threshold_support} were empirically determined through a multi-faceted analysis. First, we identified calibration-consistent regions in Fig.~\ref{fig:calibration_curves} where predicted probabilities align with observed frequencies. Second, we analyzed precision-recall trade-offs using confusion matrix metrics to balance safety-critical detection rates with false alarm minimization. Finally, we incorporated human observation to ensure practical relevance for rider assistance systems. This integrated approach ensures the thresholds are statistically supported and reliable for both baseline monitoring and safety-critical ITS decision-making. Grounded in threshold calibration theory~\cite{rajaraman2022deep,10.5555/3618408.3619385}, this framework leverages calibrated probabilities for robust risk stratification. For PTWs, where perception and urgency strongly shape warning effectiveness, these dynamically calibrated thresholds enhance both safety and interpretability~\cite{9137689}.

\subsubsection{Mapping Predicted Probabilities to ITS Interventions}
\begin{figure*}[htbp]
\centering
\includegraphics[width=0.80\linewidth, height=6cm]{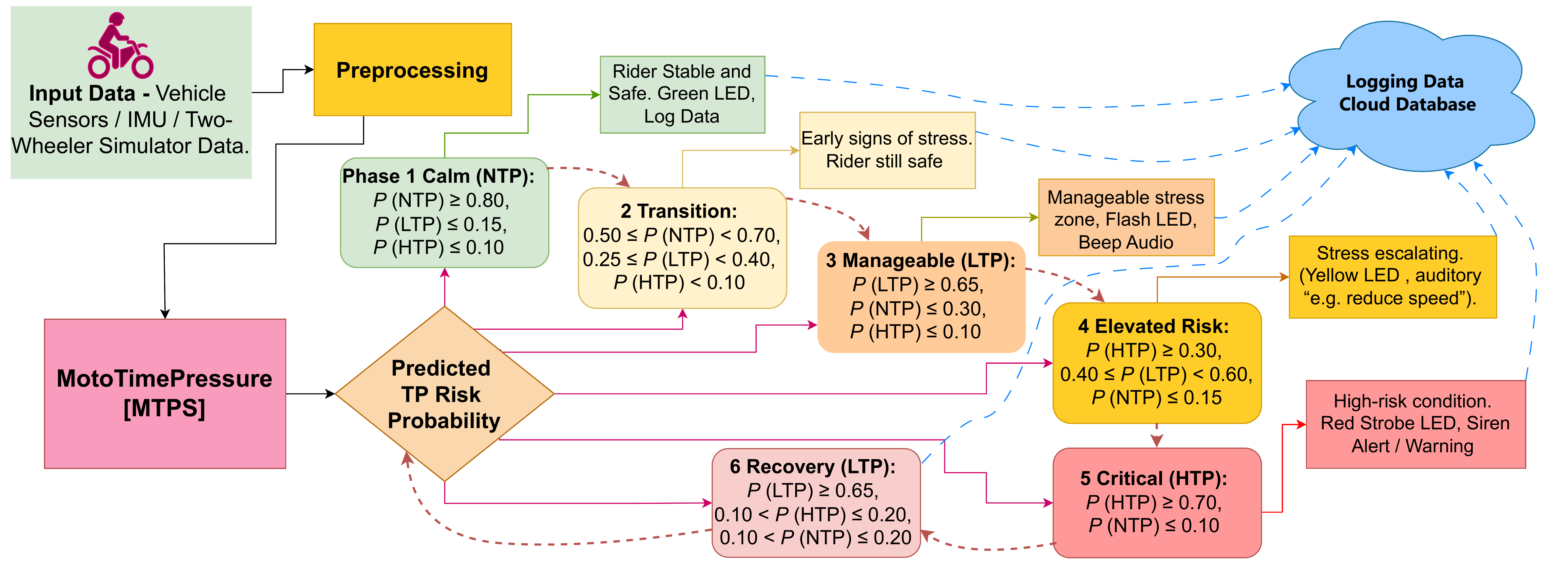}
\caption{MTPS calibrated TP prediction framework for PTW safety.}
\label{fig:MTPSPREDTPFremework}
\end{figure*}

Table~\ref{tab:threshold_support} and Fig.~\ref{fig:MTPSPREDTPFremework} show MTPS thresholds and the calibrated TP prediction framework for PTW safety.
 Predicted TP probabilities are mapped to rider states along the continuum NTP $\rightarrow$ LTP $\rightarrow$ HTP, enabling graded ITS responses. This mapping enables graded ITS interventions, from routine monitoring to critical alerts.

\subsubsection{Sensitivity and Cost Considerations}
Thresholds balance sensitivity (recall) for detecting at-risk states and precision to minimize intervention cost (e.g., unnecessary haptic alerts or throttle modulation), ensuring ITS responses are timely, reliable, and minimally disruptive.

\begin{enumerate}
    \item High thresholds (e.g., $P(\text{HTP}) \geq 0.70$, $P(\text{NTP}) \geq 0.80$) are precision-focused, minimizing false alarms and enabling strong interventions (alarms, throttle modulation).  
    \item Intermediate thresholds (e.g., $0.30 \leq P \leq 0.65$) are recall-focused, supporting early-risk monitoring and proactive alerts.  
\end{enumerate}

This trade-off ensures thresholds are statistically supported, safety-aligned, and cost-efficient, providing graded ITS responses according to rider state while minimizing unnecessary interventions.

\subsection{MTPS Model Complexity and Edge-AI Deployability}

\begin{table}[h!]
\centering
\caption{Model complexity and Edge-AI deployability of MTPS.}
\begin{tabular}{p{2cm} p{1.2cm} p{4.7cm}} 
\toprule
\textbf{Metric} & \textbf{Value} & \textbf{Positive Interpretation} \\
\midrule
Number of Parameters & 172,235 & Moderate model size; fully trainable and expressive \\
FLOPs & 0.34 M & Very low number of operations; enables efficient real-time inference \\
Inference Time & 0.21 ms (i7 CPU) & Supports sub-100ms real-time intervention on edge devices \\
Model Size & 0.66 MB & Lightweight; easy to store and deploy on most devices \\
Edge-AI Deployability & Yes & Suitable for wearable devices and on-board units \\
Deployment
Regions & Global & Can operate on wearables, on-
board units and in low-resource
settings\\
\bottomrule
\end{tabular}
\label{tab:mtps_complexity}
\end{table}

As shown in Table~\ref{tab:mtps_complexity}, MTPS is compact and computationally efficient, making it readily deployable on edge-AI platforms such as wearables and onboard units.

\section{Applications of Predicted TP}
\label{subsec:impact_tp}

\subsection{Significance of Predicted MTPS TP for Collision Prediction}
\label{subsec:impact_tp_app}

This section illustrates the significance of MTPS-predicted TP by showing how its integration as an input to the Informer substantially improves downstream collision prediction accuracy.

\subsubsection{Evaluation of MTPS-Predicted TP}
Direct measurement of GT TP labels, denoted \(\mathbf{y}\), is highly challenging in practical deployment. Therefore, it is essential to distinguish between different evaluation configurations:

\begin{enumerate}

    \item Oracle: GT labels \(\mathbf{y}\) as an additional input. Although not deployable, this establishes an upper bound on performance.  
    \item MTPS: Predicted TP labels \(\hat{\mathbf{y}}_{\text{MTPS}}\) are added as input, enabling deployable real-time inference.  
  
\end{enumerate}

The Informer model~\cite{zhou2021informer}, a Transformer-based architecture for long-sequence time-series forecasting, is used to assess the utility of MTPS-predicted TP in collision prediction. 
Given a multivariate input $X=\{x_1,\ldots,x_T\}, x_t\in\mathbb{R}^d$, it learns 
\begin{equation}
f_\text{Informer}: X \rightarrow \hat{Y}, \quad \hat{y}_t \in \mathbb{R}^C,
\end{equation}

where $\hat{\mathbf{y}}_t$ denotes class probabilities. Efficiency is achieved through \textit{ProbSparse attention}, reducing complexity from $O(L^2)$ to $O(L \log L)$ for sequence length $L$. Pooling compresses intermediate representations, and the decoder generates the full prediction horizon in a single forward pass:
\begin{equation}
\hat{Y} = \text{Decoder}(Z_\text{enc}, X_\text{start}).
\end{equation}

To quantify the importance of TP for collision prediction, the Informer model is evaluated under three input settings: 1. Baseline ($Z=X$), 2. MTPS-Predicted TP ($Z=X\oplus\hat{y}_{\text{MTPS}}$), and 3. Oracle with GT TP ($Z=X\oplus\mathbf{y}$).

\begin{table}[htbp]
\caption{Collision prediction accuracy of Informer and TimesNet with baseline, MTPS-predicted TP, and oracle.}
\label{tab:informer_results}
\centering
\begin{tabular}{lc}
\toprule
\textbf{Model and Input Configuration} & \textbf{Accuracy (\%)} \\
\midrule
\multicolumn{2}{l}{\textbf{Informer}} \\
Baseline: \( f_{\text{Informer}}(X) \) & 91.25 \\
MTPS-Predicted TP: \( f_{\text{Informer}}(X \oplus \hat{y}_{\text{MTPS}}) \) & 93.51 \\
Oracle (GT TP): \( f_{\text{Informer}}(X \oplus y) \) & 93.72 \\
\midrule
\multicolumn{2}{l}{\textbf{TimesNet}} \\
Baseline: \( f_{\text{TimesNet}}(X) \) & 92.10 \\
MTPS-Predicted TP: \( f_{\text{TimesNet}}(X \oplus \hat{y}_{\text{MTPS}}) \) & 93.90 \\
Oracle (GT TP): \( f_{\text{TimesNet}}(X \oplus y) \) & 94.06 \\
\bottomrule
\end{tabular}
\end{table}

As shown in Table~\ref{tab:informer_results}, incorporating MTPS-predicted TP into an Informer-based crash risk predictor increases accuracy from 91.25\% (baseline) to 93.51\%, approaching the oracle performance of 93.72\%. The Oracle establishes the theoretical upper bound with $\Delta_{\text{max}} = \mathcal{A}_{\text{oracle}} - \mathcal{A}_{\text{baseline}} = 2.47\%$. MTPS-predicted TP achieves $\mathcal{A}_{\text{MTPS}}=93.51\%$, yielding $\Delta_{\text{MTPS}} = \mathcal{A}_{\text{MTPS}} - \mathcal{A}_{\text{baseline}} = 2.26\%$ and a minimal gap $\epsilon = |\mathcal{A}_{\text{oracle}} - \mathcal{A}_{\text{MTPS}}| = 0.21\%$ to the Oracle ($\mathcal{A}_{\text{oracle}}=93.72\%$), corresponding to $\eta = \tfrac{\Delta_{\text{MTPS}}}{\Delta_{\text{max}}}\times 100\% \approx 91.50\%$ of the maximum benefit. TP can be interpreted as a latent variable influencing rider dynamics. The inference pipeline $\mathbf{X} \xrightarrow{f_{\text{MTPS}}} \hat{y}_{\text{MTPS}} \xrightarrow{\oplus} X \oplus \hat{y}_{\text{MTPS}} \xrightarrow{f_{\text{Informer}}} \hat{y}$ shows that predicted TP captures over $91.50\%$ of the ground-truth benefit. The generalizability of this finding is further supported by TimesNet~\cite{wu2023timesnettemporal2dvariationmodeling}, which achieved 94.06\% accuracy with ground truth TP (93.90\% with MTPS-predicted TP), compared to 92.10\% without TP. This demonstrates that TP effectively reflects rider cognitive stress and control variability, serving as a discriminative, interpretable feature for real-time ITS interventions.

\subsection{Application in ITS Interventions}

The predicted TP probabilities are thresholded to classify riders into discrete states (NTP, LTP, HTP), as shown in Table~\ref{tab:threshold_support} and Fig.~\ref{fig:MTPSPREDTPFremework}. This state-based risk index reframes ITS from passive monitoring to proactive, preventive interventions. Under the SSA, the emphasis is not only on preventing crashes but also on minimizing harm when crashes occur. By integrating MTPS-predicted TP states into adaptive human–machine interfaces (haptics, audio, AR) and vehicle-to-infrastructure (V2I) signaling (e.g., road side unit-based speed guidance, dynamic signal phasing, or targeted alerts), interventions can be triggered before risky behavior escalates. This enables systems to modulate kinetic energy exposure in real time (e.g., slowing riders earlier, guiding safer gap acceptance, or synchronizing vehicle–infrastructure interactions) thereby reducing the severity of crashes even when they are unavoidable. MTPS thus translates cognitive stress estimation into actionable intelligence for SSA-compliant ITS, where thresholds are aligned with both crash likelihood and energy transfer risk, supporting holistic harm reduction strategies.

\section{Discussions and Conclusions}

This study introduces the first dedicated framework for cognitive state prediction in PTW riders, including a dedicated TP dataset comprising 129,209 feature windows extracted via sliding window segmentation from 51 riders across 153 simulator sessions (NTP, LTP, HTP). Our empirical analysis reveals systematic behavioral degradation under TP, with HTP conditions exhibiting 48\% higher speeds and compound risks across multiple riding domains including 58\% more risky turns, 36\% more sudden braking, and 67\% worse clutch control. These findings establish the critical need for automated TP detection, as human compensation alone proves insufficient to eliminate collision risks: despite natural adaptation behaviors, TP still causes 21.3\% worse dangerous control and 10.2\% more severe violations.

Building on these empirical insights that demonstrate the systemic impact of TP on riding safety, we developed MTPS, a DL architecture that achieves 91.53\% accuracy and 98.93\% ROC AUC in classifying TP states, outperforming six baselines with only 172K parameters, 0.66 MB model size, and 0.21 ms inference on CPU. Beyond classification accuracy, MTPS demonstrates significant practical utility by improving downstream crash-risk prediction. Using MTPS-predicted TP as a feature improves collision risk accuracy for both Informer (91.25\% to 93.51\%) and TimesNet (92.10\% to 93.90\%), approaching oracle performance (93.72\% and 94.06\%, respectively). Additional validation confirms perfect HTP discrimination (AUC = 1.00) and reliable confidence calibration essential for real-world deployment. MTPS operates on sensor data from individual riders, making it adaptive to rider-specific behavior without assuming homogeneity. Additionally, TP manifests through universal kinematic signals—increased speed, sudden braking, and rapid acceleration—that are consistent across driving environments.

In conclusion, MTPS is a compact, interpretable TP prediction framework (172K parameters, 0.34M FLOPs, 0.66 MB) that can be deployed on edge-AI platforms such as wearables and on-board units for real-time monitoring. By operationalizing SSA and kinetic energy principles in ITS interventions, MTPS bridges methodological innovation and policy relevance, advancing the design of cognitively adaptive and harm-aware mobility systems. Future work will validate MTPS in naturalistic riding, extend to multimodal signals, and test adaptive thresholding for real-world ITS interventions.

\subsection{Limitations and Future Work}

This study has several limitations. First, our sample includes only male participants, reflecting Indian demographic patterns where males account for 85.2--87.3\% of PTW fatalities, but this limits the generalizability to female riders, as TP perception and riding behavior may differ by gender. Second, data were collected in a controlled simulator environment, which may not fully capture real-world dynamics such as fatigue, weather, or social pressures. Collecting real-world TP data is challenging due to safety and ethical constraints. To address this, we plan to collaborate with commercial riding platforms to collect naturalistic riding data and validate MTPS in real traffic scenarios. 

Future research will validate MTPS on large-scale simulator and naturalistic data across diverse populations and road environments. Multimodal signals (physiological, behavioral, contextual) will be integrated to strengthen robustness, while adaptive ITS interventions (haptics, throttle modulation, context-aware alerts) will be evaluated for their ability to continuously regulate both crash risk and kinetic energy transfer. Transfer learning and domain adaptation will support scalability across vehicle types, regions, and cultures. Ultimately, MTPS provides a pathway to operationalize the SSA principles (accepting that human error is inevitable but preventable harm is not) by embedding proactive cognitive load management into the design of next-generation ITS.

\section*{Funding}
This work was supported by the IIT Indore Young Faculty Research Catalyzing Grant (YFRCG) Scheme [Project ID: IITI/YFRCG/2023-24/01].

\section*{Acknowledgments}
The authors sincerely thank all volunteers who participated in this study. Special thanks to Manvendra T. for his support in data collection.

\section*{CRediT authorship contribution statement}

\textbf{Sumit S. Shevtekar:} Conceptualization, Methodology, Data curation, Formal analysis, Software, Visualization, Writing – original draft. \textbf{Chandresh K. Maurya:} Supervision, Methodology, Conceptualization, Validation, Data curation, Writing – review \& editing.
\textbf{Gourab Sil:} Supervision, Methodology, Conceptualization, Data curation, Writing – review \& editing.

\section*{Data Availability}
All data and code used will be made available from the corresponding author on a reasonable request upon publication.

\section*{Competing Interests}
The authors declare that they have no known competing financial interests or personal relationships that could have appeared to influence the work reported in this paper.

\bibliographystyle{IEEEtran}
\bibliography{references}

\begin{IEEEbiography}[{\includegraphics[width=1in,height=1.15in,clip,keepaspectratio]{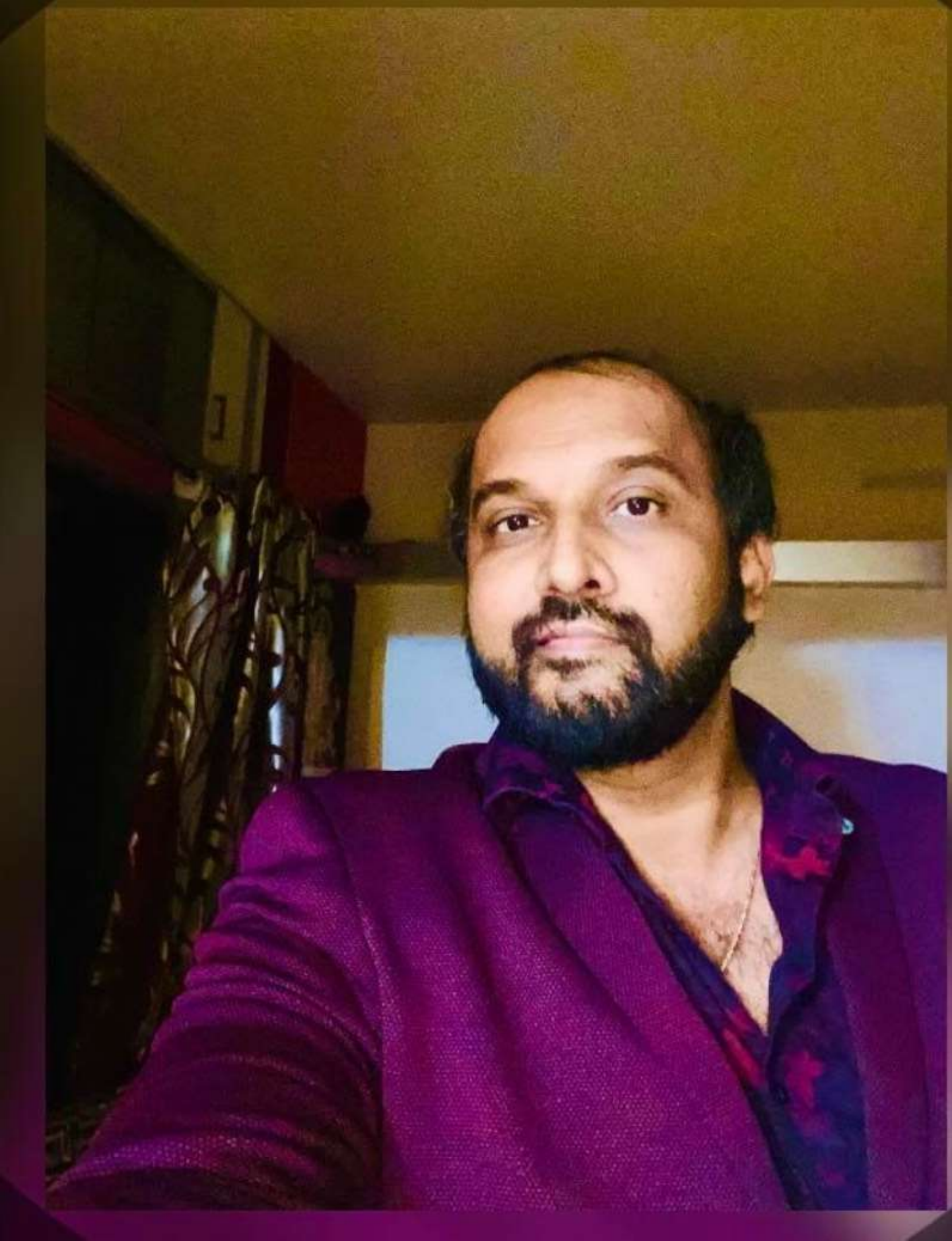}}]{Sumit S. Shevtekar}
Sumit S. Shevtekar is a Ph.D. scholar in Computer Science and Engineering at IIT Indore and Assistant Professor at PICT Pune. His research, in collaboration with Texas State University, focuses on AI and machine learning for two-wheeler safety, driver behavior analysis, ITS, and AI applications in transportation. He holds B.E. and M.E. degrees in Computer Engineering from SIT Lonavala, Pune University.
\end{IEEEbiography}

\vspace{-30pt}
\begin{IEEEbiography}[{\includegraphics[width=1in,height=1.15in,clip,keepaspectratio]{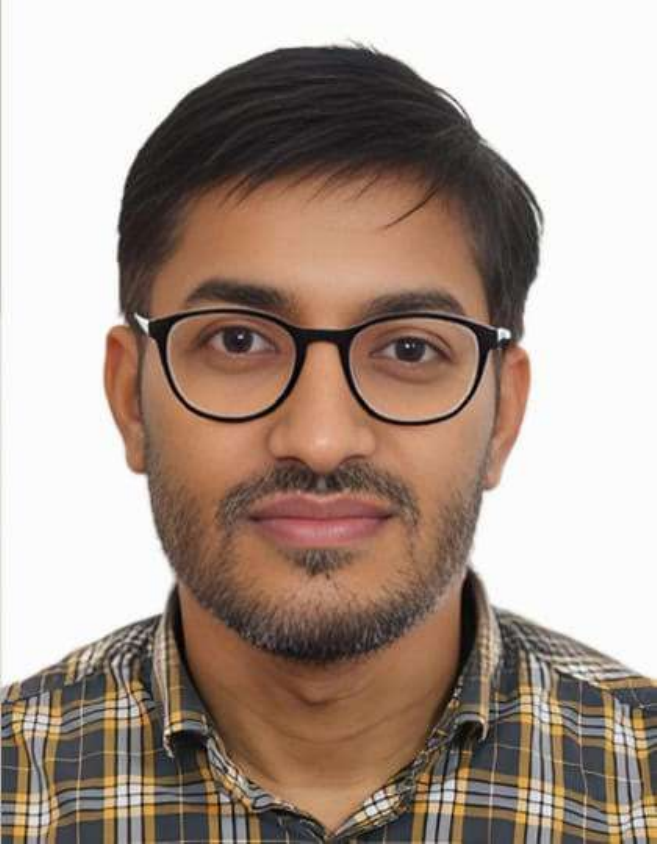}}]{Chandresh K. Maurya}
Dr. Chandresh K. Maurya is Associate Professor in Computer Science and Engineering at IIT Indore. He earned his Ph.D. from IIT Roorkee (2016, Prime Minister Fellowship), and worked at IBM Research India and ELTE University, Budapest. His research focuses on deep learning and ML for two-wheeler safety, driver behavior, ITS, and smart urban mobility.
\end{IEEEbiography}

\vspace{-30pt}

\begin{IEEEbiography}[{\includegraphics[width=1in,height=1.15in,clip,keepaspectratio]{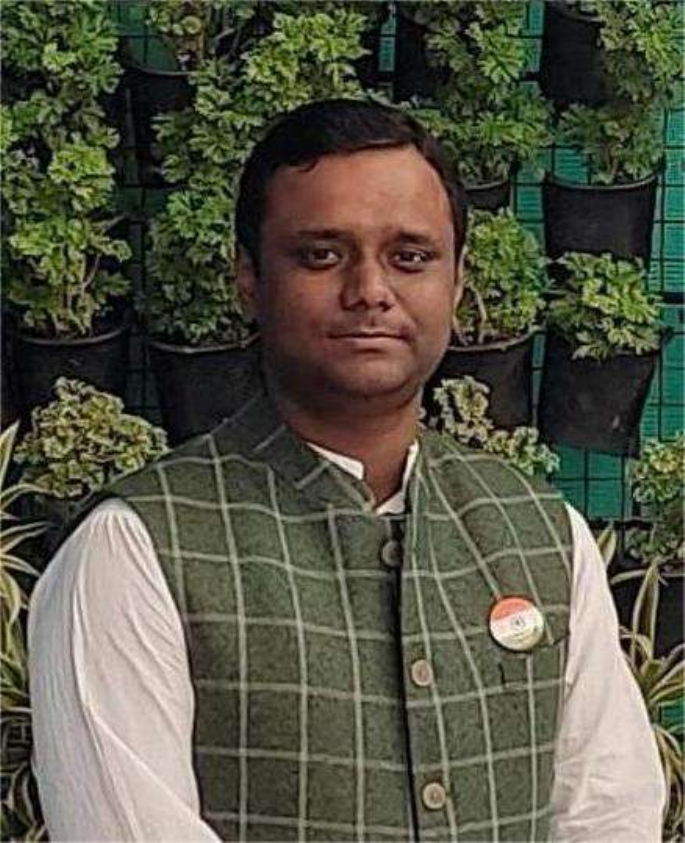}}]{Gourab Sil}
Dr. Gourab Sil is Head of the Department and Assistant Professor in Civil Engineering at IIT Indore. He earned his Ph.D. in Transportation Systems Engineering from IIT Bombay (2019), previously served as Visiting Faculty at BITS Pilani and Research Associate at IIT Bombay. His research focuses on ITS, geometric highway design, transportation safety, and driver behavior. 
\end{IEEEbiography}

\vspace{-30pt}

\vfill

\end{document}